\documentclass[lettersize,journal]{IEEEtran}
\usepackage{amsmath,amsfonts}
\usepackage{algorithmic}
\usepackage{array}
\usepackage[caption=false,font=normalsize,labelfont=sf,textfont=sf]{subfig}
\usepackage{textcomp}
\usepackage{stfloats}
\usepackage{url}
\usepackage{verbatim}
\usepackage{graphicx}
\usepackage{cite}
\usepackage{amsmath,amssymb,amsfonts}             
\usepackage[ruled,linesnumbered]{algorithm2e}
\usepackage{tabularx}
\usepackage{multirow}
\usepackage{booktabs}                                          

\def\BibTeX{{\rm B\kern-.05em{\sc i\kern-.025em b}\kern-.08em
		T\kern-.1667em\lower.7ex\hbox{E}\kern-.125emX}}
\begin{document}
	
	\title{Adversarial Federated Consensus Learning \\for Surface Defect Classification \\Under Data Heterogeneity in IIoT}
	\author{
		Jixuan Cui,~Jun Li,~Zhen Mei,~Yiyang Ni,~Wen Chen,~Zengxiang Li

		\thanks{Jixuan Cui, and Zhen Mei are with the School of Electronic and Optical Engineering, Nanjing University of Science and Technology, Nanjing 210094, China (e-mail: jixuancui@njust.edu.cn; meizhen@njust.edu.cn).}
		\thanks{Jun Li is with the School of Information Science and Engineering, Southeast University, Nanjing 210096, China (e-mail: jleesr80@gmail.com).}
		\thanks{Yiyang Ni is with the School of Computer Engineering, Jiangsu Second Normal University, Nanjing 210013, China, and also with Jiangsu Key Laboratory of Wireless Communications, Nanjing University of Posts and Telecommunications, Nanjing 210003, China (e-mail: niyy@jssnu.edu.cn).}
		\thanks{Wen Chen is with the Department of Electronics Engineering, Shanghai Jiao Tong University, Shanghai 200240, China (e-mail: wenchen@sjtu.edu.cn).}
		\thanks{Zengxiang Li is with the Digital Research Institute, ENN Group, Langfang 065001, China (e-mail: lizengxiang@enn.cn).}
	}

	\markboth{}  
	{Shell \MakeLowercase{\textit{et al.}}: A Sample Article Using IEEEtran.cls for IEEE Journals}
	
	\maketitle
	
	\vspace{-5cm}
	
	\begin{abstract}
		The challenge of data scarcity hinders the application of deep learning in industrial surface defect classification (SDC), as it's difficult to collect and centralize sufficient training data from various entities in Industrial Internet of Things (IIoT) due to privacy concerns. Federated learning (FL) provides a solution by enabling collaborative global model training across clients while maintaining privacy. However, performance may suffer due to data heterogeneity—discrepancies in data distributions among clients. In this paper, we propose a novel personalized FL (PFL) approach, named Adversarial Federated Consensus Learning (AFedCL), for the challenge of data heterogeneity across different clients in SDC. First, we develop a dynamic consensus construction strategy to mitigate the performance degradation caused by data heterogeneity. Through adversarial training, local models from different clients utilize the global model as a bridge to achieve distribution alignment, alleviating the problem of global knowledge forgetting. Complementing this strategy, we propose a consensus-aware aggregation mechanism. It assigns aggregation weights to different clients based on their efficacy in global knowledge learning, thereby enhancing the global model's generalization capabilities. Finally, we design an adaptive feature fusion module to further enhance global knowledge utilization efficiency. Personalized fusion weights are gradually adjusted for each client to optimally balance global and local features. Compared with state-of-the-art FL methods like FedALA, the proposed AFedCL method achieves an accuracy increase of up to 5.67\% on three SDC datasets.
	\end{abstract}
	
	
	%【1】 
	\section{Introduction}
	
	\IEEEPARstart{I}{ndustrial} Internet of Things (IIoT) has been widely deployed across various industrial sectors, facilitating distributed data acquisition and real-time monitoring among geographically dispersed industrial entities. Surface defect inspection, as one of the critical tasks of IIoT, has gained significant attention due to its impact on the quality and safety of industrial products. In recent decades, substantial research has focused on developing methods to detect and classify surface defects, thereby enabling production line adjustments and quality improvements\iffalse[1]-[6]\fi\cite{8948233,9220110, 9049117,10463617,10243621, 10549926}. Traditional surface defect classification (SDC) methods often rely on manual inspection, which is labor-intensive, time-consuming, and susceptible to human errors. \par
	
	In recent years, the integration of artificial intelligence with IIoT has provided new automation solutions across various industries, significantly enhancing efficiency and effectiveness\iffalse[7]-[14] \fi \cite{CHU2017140, CHU2018108, DBLP:journals/mta/WangXYLS18, 8421627, 8657760, met11030388, lin2022small, 9044434}. Traditional machine learning classifiers, such as support vector machine (SVM) and random forest \iffalse[7]-[9]\fi\cite{CHU2017140, CHU2018108,  DBLP:journals/mta/WangXYLS18}, have been utilized for automated SDC. These classifiers are typically coupled with some shape-based and texture-based feature extraction methods \iffalse[10]\fi\cite{8421627}, such as Fourier descriptor and local binary pattern. However, they suffer from limited feature extraction capabilities due to reliance on manual feature engineering, inadequately capturing the diversity of surface defects, thus limiting generalization and adaptability. On the contrary, deep learning (DL)-based methods are capable of learning abstract representations autonomously, without the need for manual feature engineering. In particular, DL-based classifiers built on convolutional neural network (CNN) architectures, such as ResNet and MobileNet, consistently demonstrate superior performance in SDC tasks \iffalse[11-13]\fi\cite{8657760, met11030388, lin2022small}. \par
	
	Nonetheless, DL-based SDC methods require sufficient data to develop robust models, which a single entity often cannot provide \iffalse[15]\fi\cite{8764994}. To overcome this limitation, data must be collected from multiple entities, and aggregated at a centralized location, such as a central server. This process of data aggregation not only exposes sensitive information to potential breaches but also conflicts with privacy regulations.
	Federated learning (FL), a paradigm enabling model training across distributed participants, known as clients, while preserving data privacy and security, effectively addresses the aforementioned challenge of data scarcity and privacy \iffalse[16]-[25]\fi\cite{9709603,10555290,10258027,10.1145/3298981, 9664296, 9833437, 10251703, ye2023heterogeneous, pmlr-v54-mcmahan17a, MLSYS2020_38af8613}. By allowing data to remain local and only merging model parameters in the server, FL effectively enhances DL model performance without compromising data privacy. \par
	
	However, the practical implementation of FL encounters significant challenges due to data heterogeneity. Variations in production conditions, such as differences in equipment and environmental factors, result in non-independent and identically distributed (non-IID) data across participating clients. For instance, different industrial entities may encounter different categories of defect samples, leading to discrepancies in the feature representations learned by their models. Such discrepancies make it difficult for different clients to reach a consensus on learned representations. When the uploaded models are aggregated, the global model may become biased towards certain clients, thereby undermining overall performance \iffalse[23]\fi\cite{ye2023heterogeneous}. \par
	
	Here, we summarize the two key challenges in real-world SDC scenarios. First, data scarcity and privacy concerns significantly impair the capacity of industrial entities to gather sufficient data essential for training DL models. Second, variations in production conditions across different industrial entities result in non-IID data. Given that most learning algorithms assume data to be IID, DL models may suffer significant performance degradation due to the data heterogeneity among different clients in IIoT. \par
	
	To address the aforementioned challenges, we propose a personalized FL (PFL) approach, named Adversarial Federated Consensus Learning (AFedCL). It facilitates collaborative training among heterogeneous clients, developing personalized models tailored to each client's specific data distribution while preserving privacy. Distinguishing from prior PFL algorithms \iffalse[26]-[30]\fi\cite{arivazhagan2019federated, collins2021exploiting, li2021ditto, zhang2023fedala, 10177379}, AFedCL focuses on the issue of global knowledge forgetting in local training due to distinct data distribution. Through adversarial training, it continuously evaluates the effectiveness of global knowledge learning and updates the model accordingly. Based on this strategy of retaining global knowledge locally for each client, AFedCL aligns the distributions across heterogeneous clients, effectively mitigating performance degradation caused by data heterogeneity. \par
	
	The main contributions of this paper are as follows: \par
	\begin{itemize}    
		\item[1)]
		We propose AFedCL, a new PFL method designed to tackle the challenge of data heterogeneity across different clients in SDC. To the best of our knowledge, this marks the first effort to employ FL in overcoming data scarcity and privacy concerns in SDC.
	\end{itemize}
	\begin{itemize}    
		\item[2)]
		We develop a dynamic consensus construction strategy to mitigate the performance degradation caused by data heterogeneity. By utilizing adversarial training and leveraging the global model as a bridge, this strategy aligns client data distributions and alleviates global knowledge forgetting during local training.
	\end{itemize}
	\begin{itemize}    
		\item[3)]
		We extend the dynamic consensus construction strategy with a consensus-aware aggregation mechanism. It assigns aggregating weights based on each client's global knowledge learning efficacy, measured by discrimination loss, enhancing the global model's generalization ability.
	\end{itemize}
	\begin{itemize}    
		\item[4)]
		We design an adaptive feature fusion module to further improve the efficiency of global knowledge utilization. This module tunes a fusion weight combining global and local features for each client, thereby achieving the optimal balance between global and local features.
	\end{itemize}
	
	The proposed method achieves superior results in extensive experiments on three strip steel SDC datasets. Compared with state-of-the-art methods like FedALA \cite{zhang2023fedala}, the proposed AFedCL method increases accuracy by up to 5.67\%. The rest of this paper is organized as follows. The related work is introduced in Section \MakeUppercase{\romannumeral 2}, and the FL problem in SDC is formulated in Section \MakeUppercase{\romannumeral 3}. Section \MakeUppercase{\romannumeral 4} presents the proposed method in detail. Numerical experiments are conducted in Section \MakeUppercase{\romannumeral 5} to verify the effectiveness of the proposed AFedCL framework, and Section \MakeUppercase{\romannumeral 6} concludes this paper.

	% 【2】 
	\section{Related Works}
	\subsection{Surface Defect Classification under Data Scarcity}
	Previous studies have addressed the issue of data scarcity through two main aspects. On the one hand, some research focuses on leveraging few-shot learning algorithms. For instance, Kim et al. \iffalse[15]\fi\cite{8764994} proposed a siamese neural network with CNN structure to overcome the data scarcity challenge in steel SDC. Similarly, Xiao et al. \iffalse[31]\fi\cite{9761830} applied graph embedding and distribution transformation modules, coupled with optimal transport for transductive classification. On the other hand, an alternative strategy involves the use of a generative adversarial network (GAN) to synthesize defect samples, thus enlarging the sample volume. In \iffalse[32]\fi\cite{jain2022synthetic}, three GANs are trained to augment the data for surface defects. Furthermore, a cycle-consistent adversarial network with attention mechanism, named AttenCGAN, was proposed to address the challenges of small intra-class differences and data scarcity \iffalse[33]\fi\cite{9760037}. \par
	
	Despite some progress in overcoming the challenges posed by limited data, the performance of these models is still held back by the small size of local datasets. In the scenario of distributed industrial data sources, the efficient exploitation of distributed data sources, while maintaining privacy, can significantly boost model performance and unlock the full potential of the data. \par

	\begin{figure}[t]
	\centering
	\vspace{0.4cm}
	\includegraphics[width=1\columnwidth]{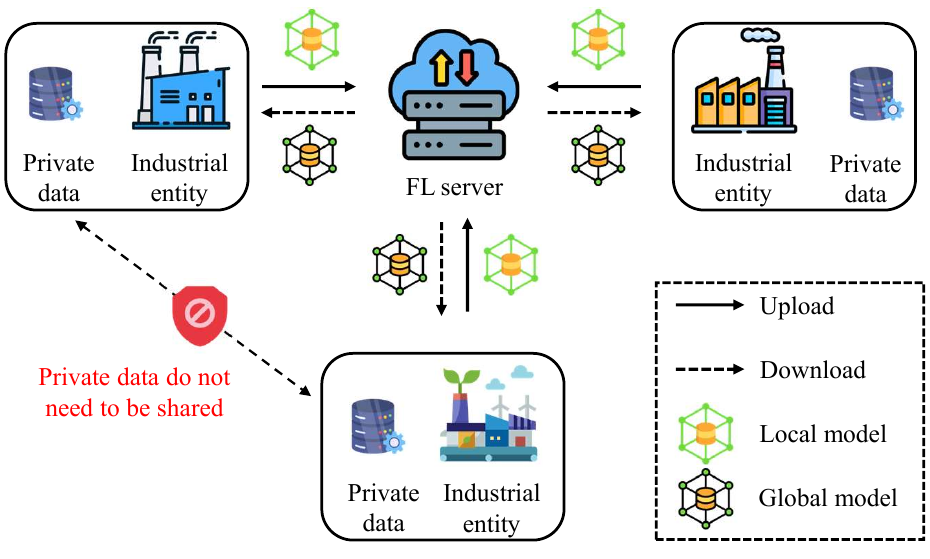}
	\vspace{0.2cm}
	\caption{Illustration of the FL architecture for IIoT scenarios. }
	\label{fig.1}
\end{figure}
	
	\subsection{Federated Learning for Data Heterogeneity}
	As shown in Fig. 1, FL aims to attain an optimized global model in the central server by aggregating local models from distributed clients, such as industrial entities, without compromising their data privacy. FedAvg \iffalse[24]\fi\cite{pmlr-v54-mcmahan17a}, recognized as the pioneering method in FL, has received increasing attention in recent years. To overcome the challenge of data heterogeneity, FedProx \iffalse[25]\fi\cite{MLSYS2020_38af8613} added a proximal term on FedAvg, constraining the discrepancies between global and local models. \par

	As the field evolved, the focus shifted towards developing PFL algorithms to cope with the problem of data heterogeneity. These algorithms aim to train personalized models for each participating client rather than a ``one-fit-all" global model. FedPer \iffalse[26]\fi\cite{arivazhagan2019federated} and FedRep \iffalse[27]\fi\cite{collins2021exploiting} adopted parameters decoupling strategy, dividing the model into shared layers and private layers, where the private layers are trained locally and not shared with the server. Ditto \iffalse[28]\fi\cite{li2021ditto} leveraged multi-task learning to learn personalized local models. FedALA \iffalse[29]\fi\cite{zhang2023fedala} adaptively aggregated the downloaded global model and local model towards the local objective on each client to initialize the local model before training in each iteration. \par
	
	Previous FL algorithms designed for heterogeneous data have made significant efforts to integrate global knowledge into local training. This integration enables models to adapt to local datasets while concurrently assimilating comprehensive global knowledge. Nevertheless, during the process of local training, clients with distinctive data distributions may give more precedence to local data characteristics, which could lead to the issue of global knowledge forgetting. However, there exists limited research that explicitly evaluates the learning effects of global knowledge and provides feedback. \par 
	
	In light of this, we will introduce local discriminators to dynamically evaluate the efficacy of global knowledge learning and provide real-time feedback to local training. Through adversarial iterative optimization, this methodology enables the distribution alignment among various data distributions, facilitating the dynamic construction of consensus among heterogeneous clients to avoid global knowledge forgetting.\par
	
	%【3】
	\section{Problem Formulation}
	The core problems of SDC in practical applications can be summarized as follows: \par
	
	1)	Data scarcity and privacy: 
	While the collection of raw industrial data is often straightforward, acquiring high-quality data labels is both difficult and expensive in IIoT. The data from a single entity is often insufficient for training robust DL models, and privacy issues restrict the sharing of data among entities, thus obstructing the deployment of DL models. \par
	
	2)	Data heterogeneity:
	Variations in production conditions between different entities result in non-IID data. As shown in Fig. 2, different clients hold data with distinct distributions. Since most learning algorithms assume data to be IID, model performance will be degraded due to data heterogeneity.
	\par
	\begin{figure}[t]

	\vspace{-0.2cm}
	\centering
	\includegraphics[width=0.98\columnwidth]{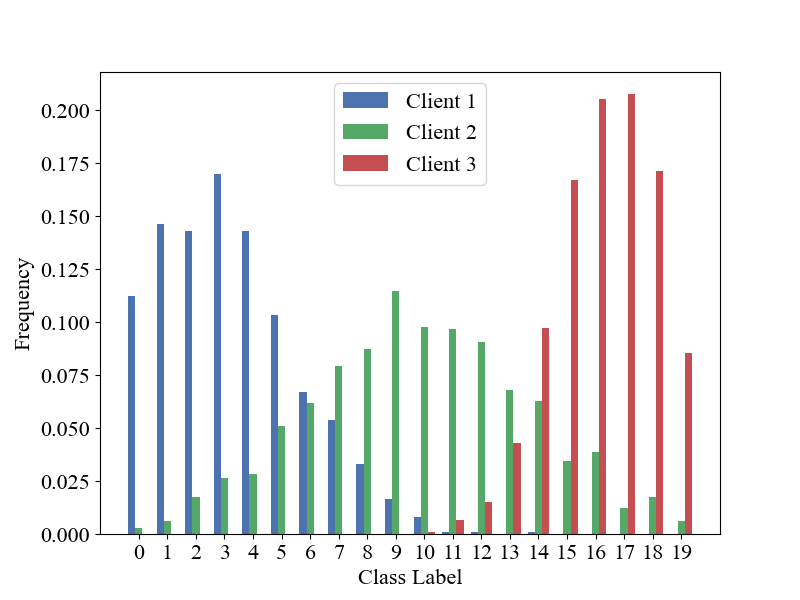}
	\caption{Illustration of non-IID data in IIoT. Each client has a distinct data distribution, which can potentially degrade the global model's performance. }
	\label{fig.2}

\end{figure}
	In the typical FL setting, a federation is established comprising a central server and a collection of $K$ participating clients. Each client $k$ possesses a private dataset $S_k = \{(x_k^l, y_k^l )\}_{l=1}^{|S_k|}$ consisting of $|S_k|$ image-label pairs, where each image $x_k^l$ and its corresponding label $y_k^l$ (indicating the class or category of the image) are sampled from a unique distribution $P_k$.
	Importantly, the dataset held by each client is relatively small, and the data distribution is unique for every client. \par
	
	Each client $k$ possesses a local model $W_k$, comprising an encoder $E_k$ and a classifier $C_k$. The encoder is responsible for extracting informative feature representations from raw images, typically employing a CNN-based backbone. Meanwhile, the classifier is tasked with categorizing the extracted feature representations to generate predictive results for the categories, often constructed using one or several fully connected layers. The empirical loss linked to model parameters $W_k$ of client $k$ and its dataset $S_k$ is denoted as $L(W_k, S_k)$, quantifying the disparity between predicted and actual values, which can be expressed as
	\begin{equation}
		L(W_k, S_k) = \frac{1}{|S_k|} \sum_{(x, y) \in S_k} l(W_k(x), y)
		\tag{1}
		\label{eq1}
	\end{equation}
	where $l(W_k(x), y)$ represents the loss for data point $(x, y)$ that penalizes the distance of model output $W_k(x)$ from label $y$ . \par
	
	The FL training process consists of multiple communication rounds, each involving parallel local training of each client and model aggregation on the server side. At each communication round, the server dispatches the global model to all clients. Each client replaces their local model with the received global model, trains it on local data, and then uploads the trained model back to the server. Then the server aggregates the received models to generate the global model for the next round. Conventionally, FL aims to learn a single global model, i.e., $W_1 = W_2 = ... = W_k = W$.  The primary objective is to minimize the aggregated empirical loss, given by
	\begin{equation}
		\min_{W}  \sum_{k=1}^{K}  p_k L(W, S_k)
		\tag{2}
		\label{eq2}
	\end{equation}
	where $p_k$ is a non-negative weight ensuring $\sum_{k} p_k = 1$. \par 
	
	However, as previously mentioned, the challenge of data heterogeneity severely deteriorates the efficacy of the global model of this overarching one-fit-all paradigm. To address this problem, PFL algorithms shift the focus from a singular, one-fit-all model to the pursuit of distinct models for each participating client. In PFL, the learning objective could be expressed as
	\begin{equation}
		\min_{W_k}  \sum_{k=1}^{K}  p_k L(W_k, S_k), \quad  k=1,2,...,K.
		\tag{3}
		\label{eq3}
	\end{equation}
	
	%【4】
	\section{Proposed Method}
	\begin{figure*}[t]
	\vspace{-0.4cm}
	\centering
	\includegraphics[width=1.88\columnwidth]{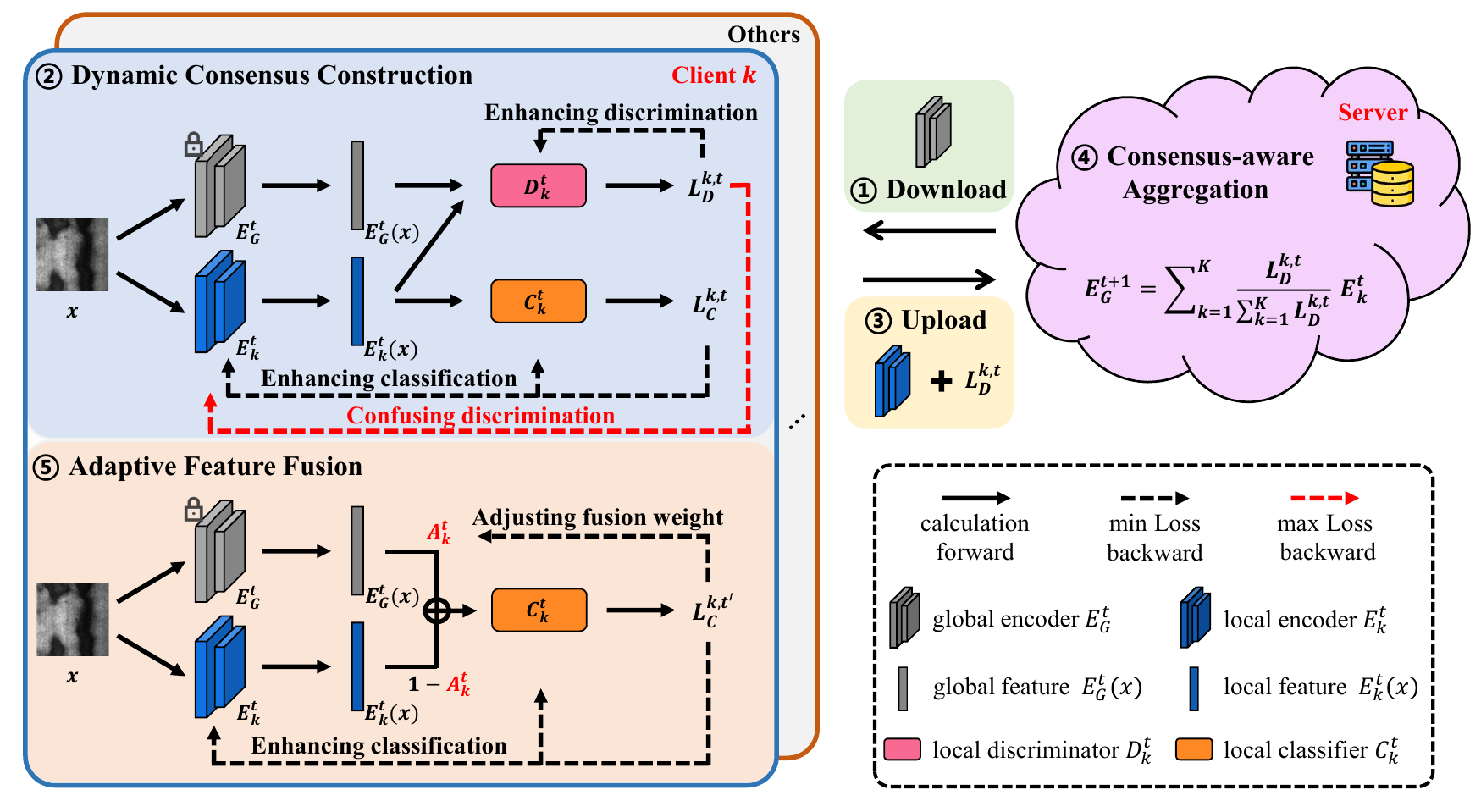}

	\caption{The workflow of the proposed AFedCL method. It comprises a central server and multiple clients, each with heterogeneous and limited training data. Through multiple communication rounds, the server and clients collaboratively train personalized models that are specifically tailored to each client’s unique local data characteristics, all while maintaining privacy and operating within the constraints of limited data.}
	\label{fig.3}
	\vspace{-0.2cm}
\end{figure*}
	\subsection{Overview} 
	This work proposes a PFL method to address data heterogeneity in SDC through three key components: a dynamic consensus construction strategy for improved global knowledge learning during local training, a consensus-aware aggregation mechanism for enhanced generalization ability of the global model, and an adaptive feature fusion module for further global knowledge utilization efficiency. Together, they collaboratively train personalized models in a privacy-preserving manner. \par
	
	As shown in Fig. 3, the overall system consists of a central server and several clients. The entire training process can be summarized into six main steps as follows: \par
	
	\begin{itemize}    
		\item[1)]Clients download the global encoder $E_G^t$ from the server; \par
	\end{itemize}
	
	\begin{itemize}    
		\item[2)]Clients update their local encoder $E_k^t$, classifier $C_k^t$, and discriminator $D_k^t$ according to the dynamic consensus construction strategy; \par
	\end{itemize}
	
	\begin{itemize}    
		\item[3)]Clients return the updated local encoder $E_k^t$ and the discrimination loss $L_D^{k,t}$ to the server; \par
	\end{itemize}
	
	\begin{itemize}    
		\item[4)]The server aggregates the encoders following the consensus-aware aggregation mechanism; \par
	\end{itemize}
	
	\begin{itemize}    
		\item[5)]Clients update local encoder $E_k^t$, classifier $C_k^t$, and fusion weight $A_k^t$ employing the adaptive feature fusion module; \par
	\end{itemize}
	
	\begin{itemize}    
		\item[6)]Repeat steps 1) through 5) until the end of training.  \par
	\end{itemize}
	
	The three components of the proposed AFedCL method will be introduced in detail.

	\subsection{Dynamic Consensus Construction}
	In scenarios where data distributions vary significantly, models may focus on learning specific characteristics of local data, risking the loss of global knowledge acquired from the global model. To ensure the persistent retention of global knowledge during local training, we develop the first stage of local training (step 2) with ongoing evaluation of global knowledge acquisition and immediate feedback based on adversarial training. Specifically, beyond the classification task, a local discriminator is trained to discriminate between the features extracted by global and local encoders from identical samples. Concurrently, the local encoder endeavors to confuse the discriminator, creating an adversarial game process between the training of the discriminator and the encoder. \par
	
	The dynamic consensus construction strategy involves the computation of two types of loss: classification loss and discrimination loss. As shown in Fig. 3, at the $t$-th communication round, client $k$ receives global encoder $E_G^t$ from the server. For a singular sample $x$, its global feature $E_G^t(x)$ and local feature $E_k^t(x)$ are obtained by inputting $x$ into the global encoder $E_G^t$ and local encoder $E_k^t$, respectively. \par
	
	For the classification loss $L_C^{k,t}$, we adopt the cross entropy loss function. The local classifier $C_k^t$ manages the local feature $E_k^t(x)$ to produce a predictive probability distribution vector $C_k^t(E_k^t(x))$. The classification loss $L_C^{k,t}$ is calculated by
	\begin{equation}
		\begin{split}
			L_C^{k,t} &= \frac{1}{|S_k|} \sum_{(x, y) \in S_k} l(C_k^t(E_k^t(x)), y) \\
			& = -\frac{1}{|S_k|} \sum_{(x, y) \in S_k} \sum_{c=1}^{N}\textbf{1}_{[y=c]}{\rm log}(C_k^t(E_k^t(x))_c) \\
		\end{split}
		\tag{4}
		\label{eq4}
	\end{equation}
	where \textbf{1} is the indicator function, and $N$ is the number of classification categories. $C_k^t(E_k^t(x))_c$ represents the probability of the $c$-th category, as predicted by the model, in $C_k^t(E_k^t(x))$. \par
	
	For the discrimination loss $L_D^{k,t}$, we still use the cross entropy loss function. However, the approach to calculate discrimination loss differs from that of classification loss in two key ways: firstly, the features fed into local discriminator $D_k^t$ now involve not only local feature $E_k^t(x)$, but also global feature $E_G^t(x)$; secondly, the category labels $\{1, 2, ..., N\}$ are substituted by domain labels $\{0, 1\}$, where $0$ for label of local features and $1$ for label of global features. Thus, the feature set for the discrimination tasks, denoted by $F_k^t$ can be defined as $F_k^t = \{(E_k^t(x),0)\}_{(x,y) \in S_k} \cup \{(E_G^t(x),1)\}_{(x,y) \in S_k}$. The discrimination loss $L_D^{k,t}$ can be calculated by 
	\begin{equation}
		\begin{split}
			L_D^{k,t} &= \frac{1}{2|S_k|} \sum_{(f, y) \in F_k^t} l(D_k^t(f), y) \\
			& = -\frac{1}{2|S_k|} \sum_{(f,y) \in F_k^t} \sum_{c=1}^{2}\textbf{1}_{[y=c]}{\rm log}(D_k^t(f)_c) \\
		\end{split}
		\tag{5}
		\label{eq5}
	\end{equation}
	where $D_k^t(f)$ is the discriminator output as a probability distribution, and $D_k^t(f)_c$ is the probability of the $c$-th category. \par
	
	Then, the classification loss $L_C^{k,t}$ is minimized to update the whole model $W_k^t$ including local encoder $E_k^t$ and classifier $C_k^t$. Meanwhile, the discrimination loss $L_D^{k,t}$ is minimized to update the discriminator $D_k^t$, and maximized to update the local encoder $E_k^t$. The model optimization problem for client $k$ at communication round $t$ can be formulated as:
	\begin{equation}
		\begin{aligned}
			\hat{E_k^t} & = arg\{ \min_{E_k^t} L_C^{k,t}, \max_{E_k^t} L_D^{k,t} \}, \\
			\hat{C_k^t} & = arg\{ \min_{C_k^t} L_C^{k,t} \}, \\
			\hat{D_k^t} & = arg\{ \min_{D_k^t} L_D^{k,t} \}. 
		\end{aligned}
		\tag{6}
		\label{eq6}
	\end{equation}
	
	At each training step, the parameters can be updated as
	\begin{equation}
		\begin{aligned}
			E_k^t & \leftarrow E_k^t - \alpha ( \frac{\partial L_C^{k,t}}{\partial E_k^t} - \lambda \frac{\partial L_D^{k,t}}{\partial E_k^t}) , \\
			C_k^t & \leftarrow C_k^t - \alpha ( \frac{\partial L_C^{k,t}}{\partial C_k^t}    ) ,\\
			D_k^t & \leftarrow D_k^t - \alpha \lambda ( \frac{\partial L_D^{k,t}}{\partial D_k^t}    ) 
		\end{aligned}
		\tag{7}
		\label{eq7}
	\end{equation}
	where $\alpha$ is the learning rate, and $\lambda$ represents the balancing weight between the discrimination loss $L_D^{k,t}$ and the classification loss $L_C^{k,t}$, which governs the relative importance of $L_D^{k,t}$ and $L_C^{k,t}$ in the optimization process. \par

	As communication rounds increase, the discriminator's ability to distinguish between global and local features strengthens, alongside the encoder's ability to confuse the discrimination. As a result, the similarity between global and local features from identical samples improves. In this way, each client with a distinct data distribution aligns its personalized knowledge with that of the global model. Consequently, the global model serves as a bridge to achieve distribution alignment among heterogeneous clients, dynamically constructing a consensus through several adversarial game processes between the clients and the server.  \par

	\begin{figure}[t]
	\centering
	\vspace{-0.3cm}
	\includegraphics[width=0.68\columnwidth]{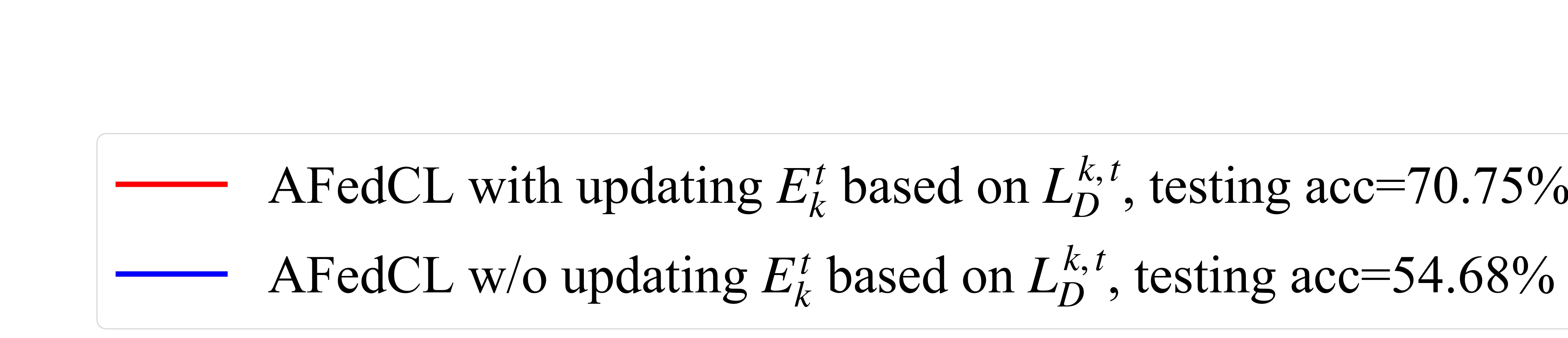} \vspace{0.0cm} \\
	
	\subfloat[]{\includegraphics[width=0.485\columnwidth]{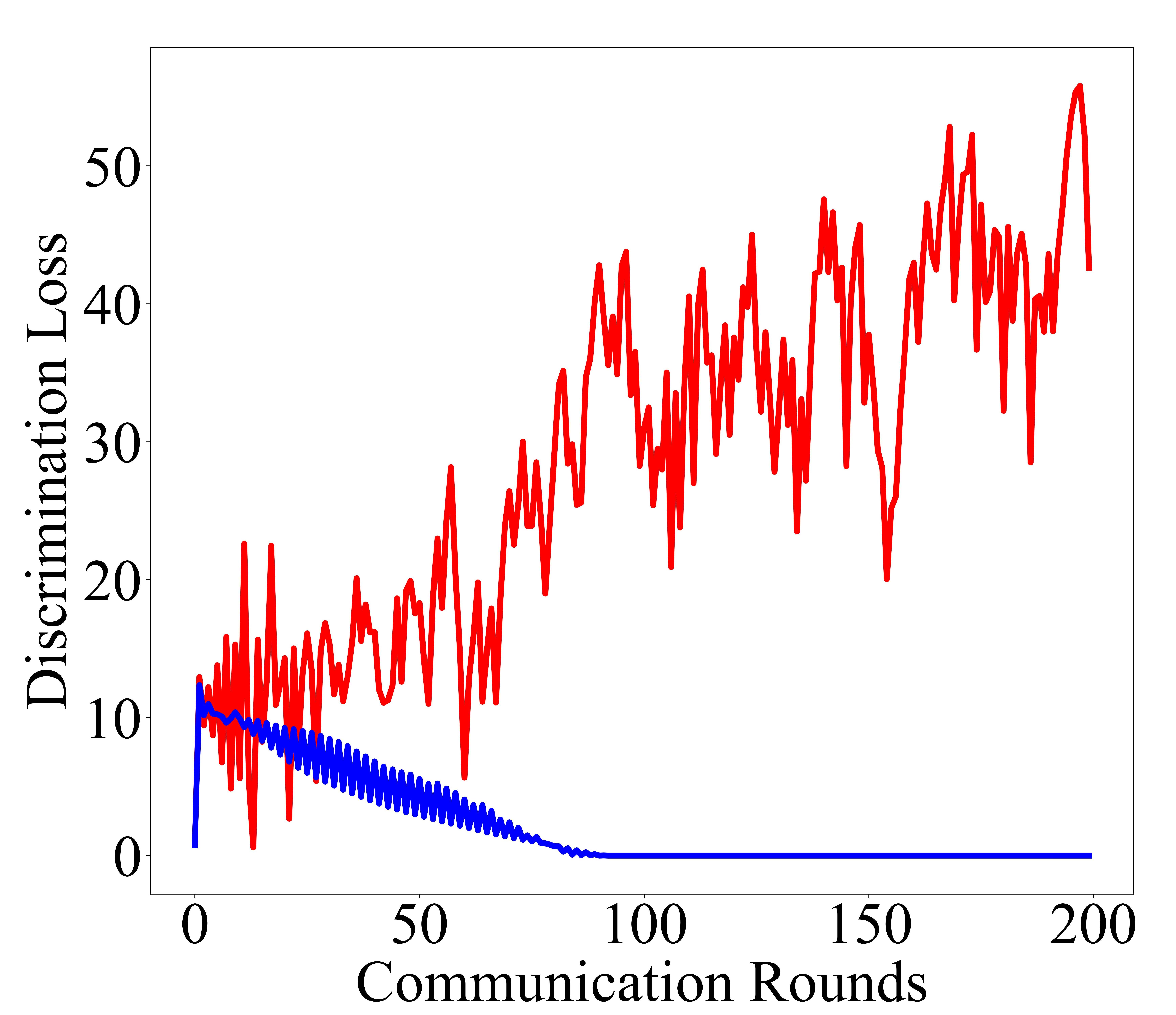}}\vspace{0.0cm}
	\subfloat[]{\includegraphics[width=0.485\columnwidth]{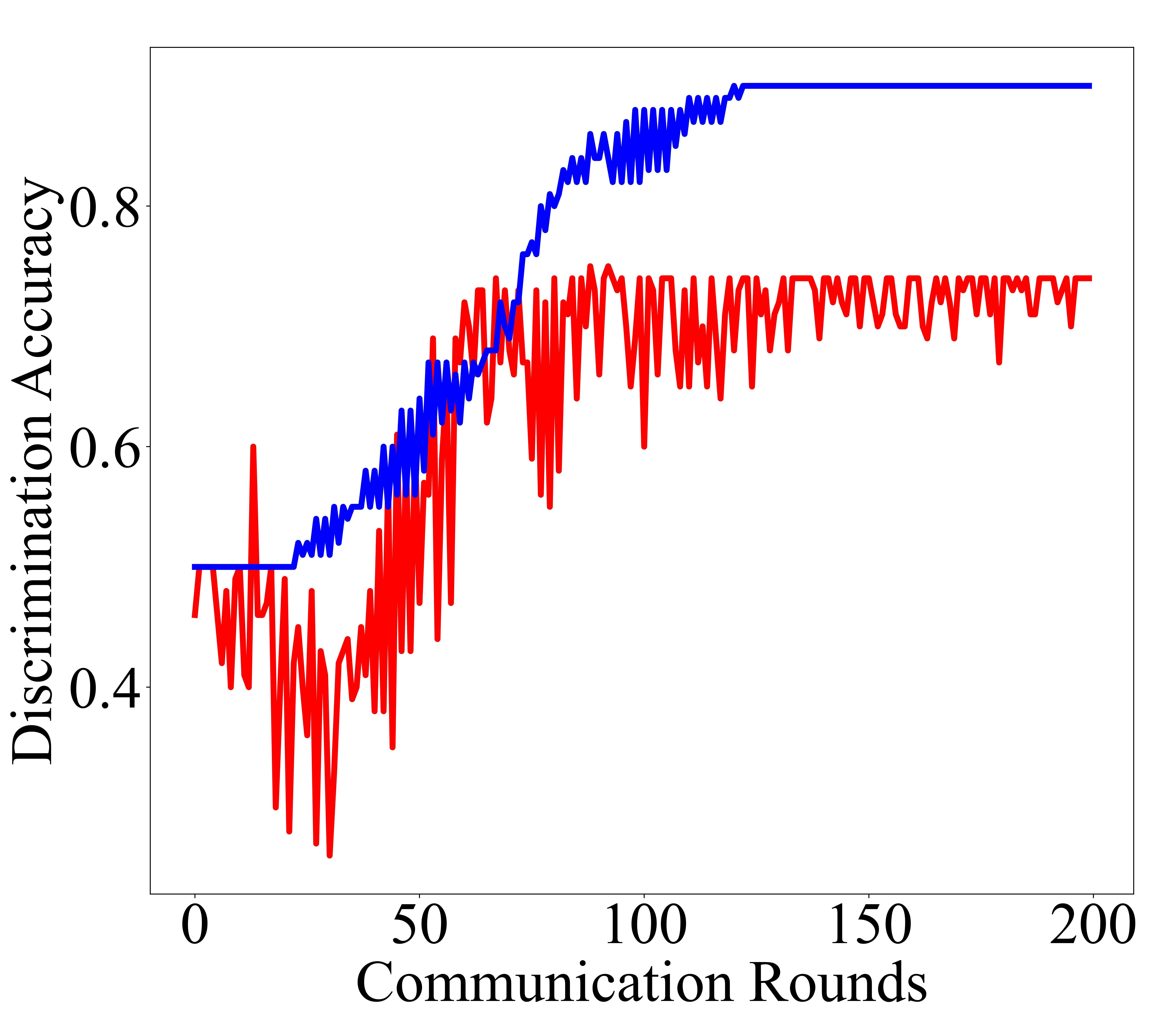} } \\
	
	\caption{Visualization of the consensus construction process. (a) discrimination loss. (b) discrimination accuracy. }
	\vspace{0.1cm}
	\label{fig.4}
\end{figure}

	\subsection{Consensus-Aware Aggregation}
	To clearly demonstrate the adversarial process and lay the groundwork for our designed consensus-aware aggregation mechanism, we conducted a sub experiment by omitting the update of $E_k^t$ based on $L_D^{k,t}$. 
	As shown in Fig. 4, without updating $E_k^t$ with $L_D^{k,t}$, discrimination loss decreases to zero, reflecting the discriminator's enhanced ability to differentiate local features from global features, as evidenced by improved discrimination accuracy.
	Conversely, updating $E_k^t$ with $L_D^{k,t}$ during dynamic consensus construction results in fluctuating discrimination loss and lower discrimination accuracy levels. 
	It reveals that the discriminator struggles to distinguish between global and local features, indicating that local encoder features closely align with global features. \par
	
	This alignment helps effectively retain global knowledge during training on private data. Reflecting these changes in final classification accuracy, the AFedCL without updating $E_k^t$ based on $L_D^{k,t}$ achieved only 54.68\%, significantly lower than the complete algorithm's 70.75\%, highlighting the essential role of the proposed strategy. \par

	Following the adversarial training at the $t$-th communication round, each client $k$ uploads $E_k^t$ and $L_D^{k,t}$ into the server for aggregation. In the server, the objective is to generate a global model with stronger generalization ability. As we discussed before, a higher discrimination loss value leads to a more similar pattern between the global and local features, indicating a greater contribution of global knowledge. Therefore, during the model aggregation stage (step 4), our designed consensus-aware aggregation mechanism assigns higher weights to clients with greater discrimination loss values. At the $t$-th round, the server aggregates models using
	\begin{equation}
		E_G^{t+1} = \sum_{k=1}^{K} \frac{L_D^{k,t}}{\sum_{k=1}^{K}L_D^{k,t}}E_k^t. \\
		\tag{8}
		\label{eq8}
	\end{equation}
	Here, $E_G^{t+1}$ is the global model for the next round. Higher aggregation weights are assigned to clients that demonstrate greater efficacy in learning global knowledge. This strategic allocation ensures the development of a global model with broader generalization capacity, rather than performing well only on the data of any specific client. \par
	
	\begin{figure}[t]
	\centering
	\includegraphics[height=0.15\textheight]{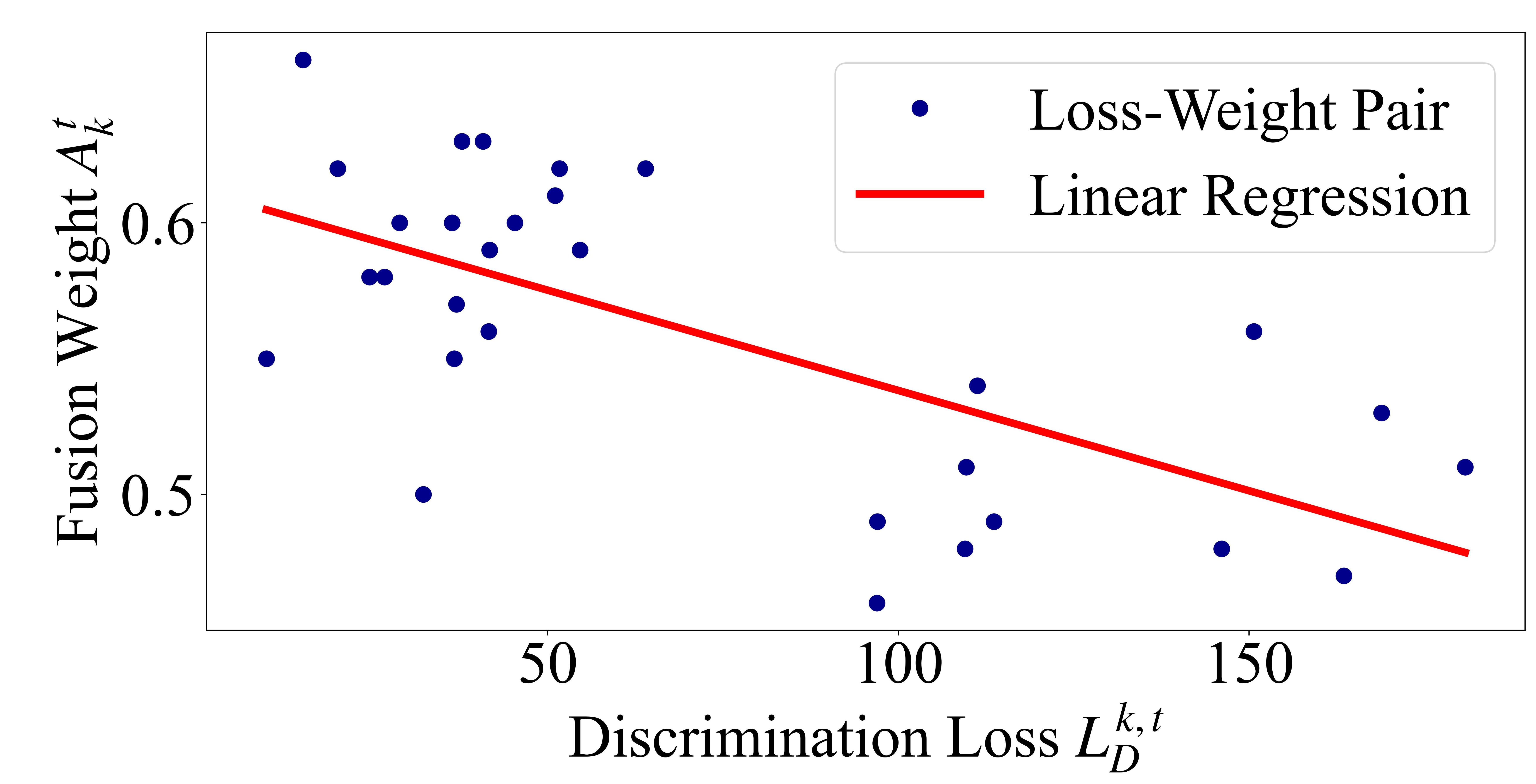}

	\caption{Relationship between discrimination loss and adaptive fusion weight.}
	\label{fig.5}
\end{figure}
	
	\subsection{Adaptive Feature Fusion}

	Considering the variations in global knowledge learning among clients, we aim to balance these differences and enhance the utilization efficiency of the global knowledge embedded in the global model. Therefore, we design an adaptive feature fusion module for the second stage of local training (step 5). 
	It gradually optimizes personalized fusion weights for each client to get an optimal balance between global and local features, combining them into more comprehensive representations for robust prediction. \par
	
	At the $t$-th communication round, the global feature $E_G^t(x)$ and the local feature $E_k^t(x)$ of sample $x$ are computed by inputting $x$ into the global encoder $E_G^t$ and the local encoder $E_k^t$. Then the adaptive fusion weight $A^k_t$ is used to generate an interpolation of them as
	\begin{equation}
		f_k^t = A_k^t \cdot E_G^t(x) + (1-A_k^t) \cdot E_k^t(x) \\
		\tag{9}
		\label{eq9}
	\end{equation}
	where $f_k^t$ is the fusion representation, and it is then fed into the local classifier $C_k^t$ to generate the final class prediction $C_k^t(f_k^t)$. The classification loss $L_C^{k,t'}$ will be computed as 
	\begin{equation}
		L_C^{k,t'} = \frac{1}{|S_k|} \sum_{(x, y) \in S_k} l(C_k^t(  A_k^t \cdot E_G^t(x) + (1-A_k^t) \cdot E_k^t(x)  ), y). \\
		\tag{10}
		\label{eq10}
	\end{equation}
	
	After that, $L_C^{k,t'}$ is minimized to update the fusion weight $A_k^t$ and the local model $W_k^t$, comprising the encoder $E_k^t$ and the classifier $C_k^t$, shown as 
	\begin{equation}
		\begin{aligned}
			W_k^t & \leftarrow W_k^t - \alpha ( \frac{\partial L_C^{k,t'}}{\partial W_k^t}   ),   \\
			A_k^t &  \leftarrow A_k^t  - \alpha ( \frac{\partial L_C^{k,t'}}{\partial A_k^t}   ).   \\
		\end{aligned}
		\tag{11}
		\label{eq11}
	\end{equation}
	
	To verify the improvements in global knowledge utilization, we analyzed data pairs including discrimination loss $L_D^{k,t}$ and fusion weight $A_k^t$ (the corresponding weight of global features) for specific clients from the final round of several experiments.
	Fig. 5 presents a linear regression analysis indicating that clients with lower discrimination loss receive larger fusion weights after training, which suggests that clients with poorer global knowledge learning efficacy exhibit a greater reliance on global features. This underscores the adaptive feature fusion module's critical role in adjusting the balance between global and local features according to each client's specific learning efficacy, further validating the use of discrimination loss as a measure of global knowledge learning efficacy in our method.

	\begin{algorithm}[t]
	
	\SetKwData{Left}{left}\SetKwData{This}{this}\SetKwData{Up}{up}
	\SetKwFunction{Union}{Union}\SetKwFunction{FindCompress}{FindCompress}
	\SetKwInOut{Input}{Input}\SetKwInOut{Output}{Output}
	
	\caption{The proposed AFedCL method. }
	\label{alg1}
	\Input{$T$, $K$, $\alpha$, $\lambda$}
	\Output{$\{  W_k^T  \}_{k=1}^{K}$}
	
	Initialize $E_G^1$, $\{  W_k^0  \}_{k=1}^{K}$, $\{  D_k^0  \}_{k=1}^{K}$, $\{  A_k^0 \}_{k=1}^{K}$

	\For{each round $t = 1, ..., T$}{
		
		Server sends the global encoder $E_G^t$ to each client\par
		
		\For{each client $k = 1, ..., K$ in parallel}{
			
			% stage 1
			Initialize models with parameters of last round: \par
			$W_k^t \leftarrow W_k^{t-1}$, $D_k^t \leftarrow D_k^{t-1}$, $A_k^t \leftarrow A_k^{t-1}$
			
			Compute $L_C^{k,t}$, and $L_D^{k,t}$ using (4)-(5)\par

			Update encoder, classifier, and discriminator: \par
			$E_k^t \leftarrow E_k^t - \alpha (\nabla_{E_k^t} L_C^{k,t} - \lambda \nabla_{E_k^t} L_D^{k,t}) $ \par
			$D_k^t \leftarrow D_k^t - \alpha \nabla_{D_k^t} L_D^{k,t}  $ \par
			$C_k^t \leftarrow C_k^t - \alpha \nabla_{C_k^t} L_C^{k,t}  $ \par
		    
		    Upload local encoder $E_k^t$ and discrimination loss $L_D^{k,t}$ to the server \par
		    
		    % stage 2
		    Compute $L_C^{k,t'}$ using (10) \par
		    Update local model and fusion weight: \par
		    $W_k^t \leftarrow W_k^t - \alpha \nabla_{W_k^t} L_C^{k,t'}  $ \par
		    $A_k^t \leftarrow A_k^t - \alpha \nabla_{A_k^t} L_C^{k,t'}  $ \par
		}
	
		Server aggregates the received models:
		$E_G^{t+1} = \sum_{k=1}^{K} \frac{L_D^{k,t}}{\sum_{k=1}^{K}L_D^{k,t}}E_k^t$ \par
	}

\end{algorithm}

	The complete training process of the proposed AFedCL method is illustrated in Algorithm 1. In practice, we implement an ``offline training-online inference" strategy. Once offline federated training is completed, clients can deploy the personalized models on the production environments. Consequently, during actual production, the deployed model can efficiently process real-time defect data and provide immediate classification predictions.\par

	% 【5】
	\section{Experiments}

	In this section, we demonstrate the necessity of adopting FL within IIoT environments and evaluate the effectiveness of the proposed AFedCL method using three real-world datasets, including two hot-rolled strip steal datasets and one cold-rolled strip steal dataset. The experiments are conducted under two non-IID settings to simulate real-world production line conditions, and we assess the method's performance under three different local training sample sizes to verify its robustness in the face of varying degrees of data scarcity.

	\subsection{Datasets} 
	\subsubsection{NEU-CLS} The Northeastern University Classification (NEU-CLS) dataset \iffalse[34] \fi\cite{9440451} is a hot-rolled strip steel SDC dataset, featuring six typical surface defect types: rolled-in scale, patches, pitted surface, crazing, inclusion, and scratches. Several typical defect images are shown in Fig. 6. \par
	\begin{figure*}[h]

	\centering
	\includegraphics[width=1.75\columnwidth]{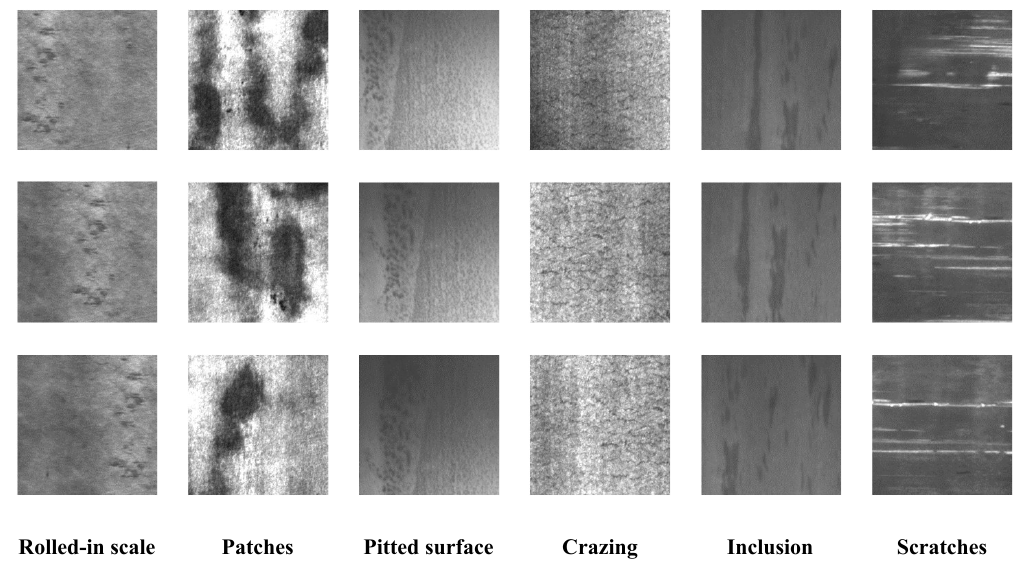}
	\caption{Typical defects of the steel surface in the NEU-CLS dataset. }
	\label{fig.6}

\end{figure*}
	\subsubsection{X-SDD} The Xsteel Surface Defect Dataset (X-SDD) \iffalse[35]\fi \cite{feng2021x}  contains seven types of hot-rolled steel strip defect images, including inclusion, red iron sheet, iron sheet ash, scratches, oxide scale of plate system, finishing roll printing, and oxide scale of the temperature system. \par
	\subsubsection{GC10-DET} The GC10-DET dataset \iffalse[36]\fi \cite{s20061562} depicts ten types of cold-rolled steel strip defect images, including punching, weld line, crescent gap, water spot, oil spot, silk spot, inclusion, rolled pit, crease, waist folding. \par

	\subsection{Experiment Setup}
	\subsubsection{Network Structure}
	In our experiments, we use the MobileNet \iffalse[37]\fi\cite{Sandler_2018_CVPR} architecture for the encoder and the classifier, and two fully connected layers for the discriminator. Moreover, in the sensitivity analysis (Section V.E), ResNet-18 and ResNet-50 \iffalse[38]\fi\cite{7780459} are employed to assess our method's performance.
	
	\subsubsection{Hyper Parameters}
	In the FL setup, we cap the communication rounds at 200. Local clients employ the Adam optimizer with a learning rate of 0.001 and a momentum coefficient of 0.9. The clients conduct 3 local epochs per round, allowing them to sufficiently adapt their models to the local data before the next communication round.

	\subsubsection{Data Distribution}
	The number of clients is set to 5. Data are distributed across two non-IID scenarios to reflect realistic challenges: 1) Disjoint: $c$ classes are randomly allocated to each client, with values of 2, 3, and 4 for the three datasets, respectively; 2) Dirichlet: data is divided using a Dirichlet distribution with $\alpha=0.1$ for all datasets. To simulate the practical scenarios, the number of training samples per client is set to 5, 10, and 20, respectively, across all datasets, reflecting varying degrees of data scarcity. Some crucial experiment parameters are shown in Table \MakeUppercase{\romannumeral 1}.
	
	\subsubsection{Baselines and Evaluation Metrics}
	We evaluate the proposed AFedCL method against several state-of-the-art baselines in three categories: 1) training on local private data without FL: ``No-FL"; 2) one-fit-all FL algorithms: FedAvg \iffalse[24]\fi\cite{pmlr-v54-mcmahan17a} and FedProx \iffalse[25]\fi\cite{MLSYS2020_38af8613}; 3) PFL algorithms: FedPer \iffalse[26]\fi\cite{arivazhagan2019federated}, FedRep \iffalse[27]\fi\cite{collins2021exploiting}, Ditto \iffalse[28]\fi\cite{li2021ditto}, and FedALA \iffalse[29]\fi\cite{zhang2023fedala}. The evaluation metrics include testing accuracy and F1 score.
	
	\begin{table}[!t]
	\caption{Experiment Parameters\label{tab:parameters}}
	\centering
	
	\begin{tabularx}{0.8\linewidth}{>{\centering\arraybackslash}m{0.4\linewidth} >{\centering\arraybackslash}m{0.4\linewidth}}
		\toprule
		Parameter & Value \\
		\midrule
		Communication rounds & 200 \\
		Optimizer & Adam \\
		Learning rate & 0.001 \\
		Momentum coefficient & 0.9 \\
		Local epochs per round & 3 \\
		Number of clients & 5 \\
		Training samples per client & 5, 10, 20 \\
		\bottomrule
	\end{tabularx}
\vspace{-0.1cm}
\end{table}

	\begin{figure*}[t]
	\centering
	\subfloat[]{\hspace{-0.3cm}\includegraphics[height=0.23\textheight]{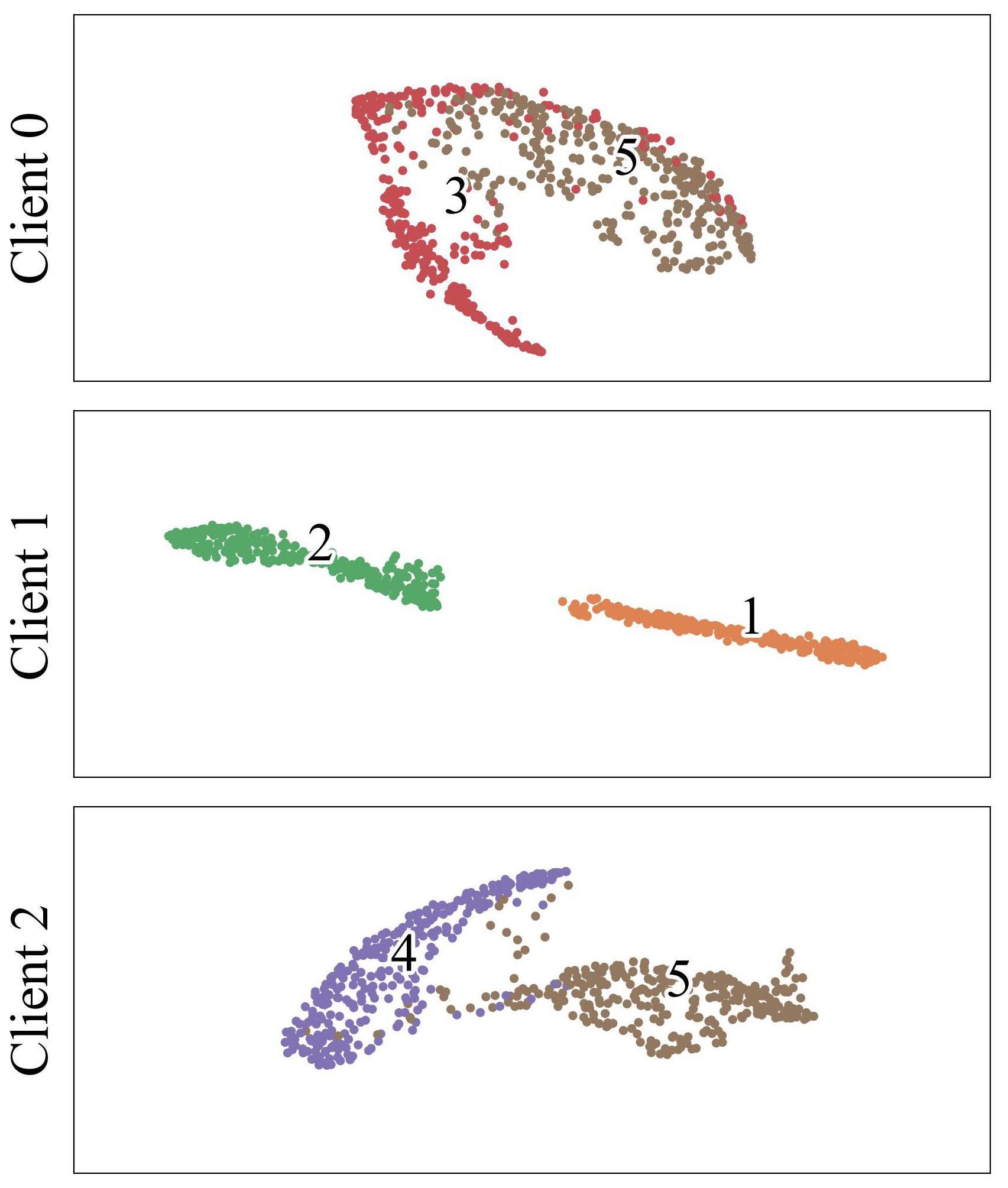}}
	\subfloat[]{\includegraphics[height=0.23\textheight]{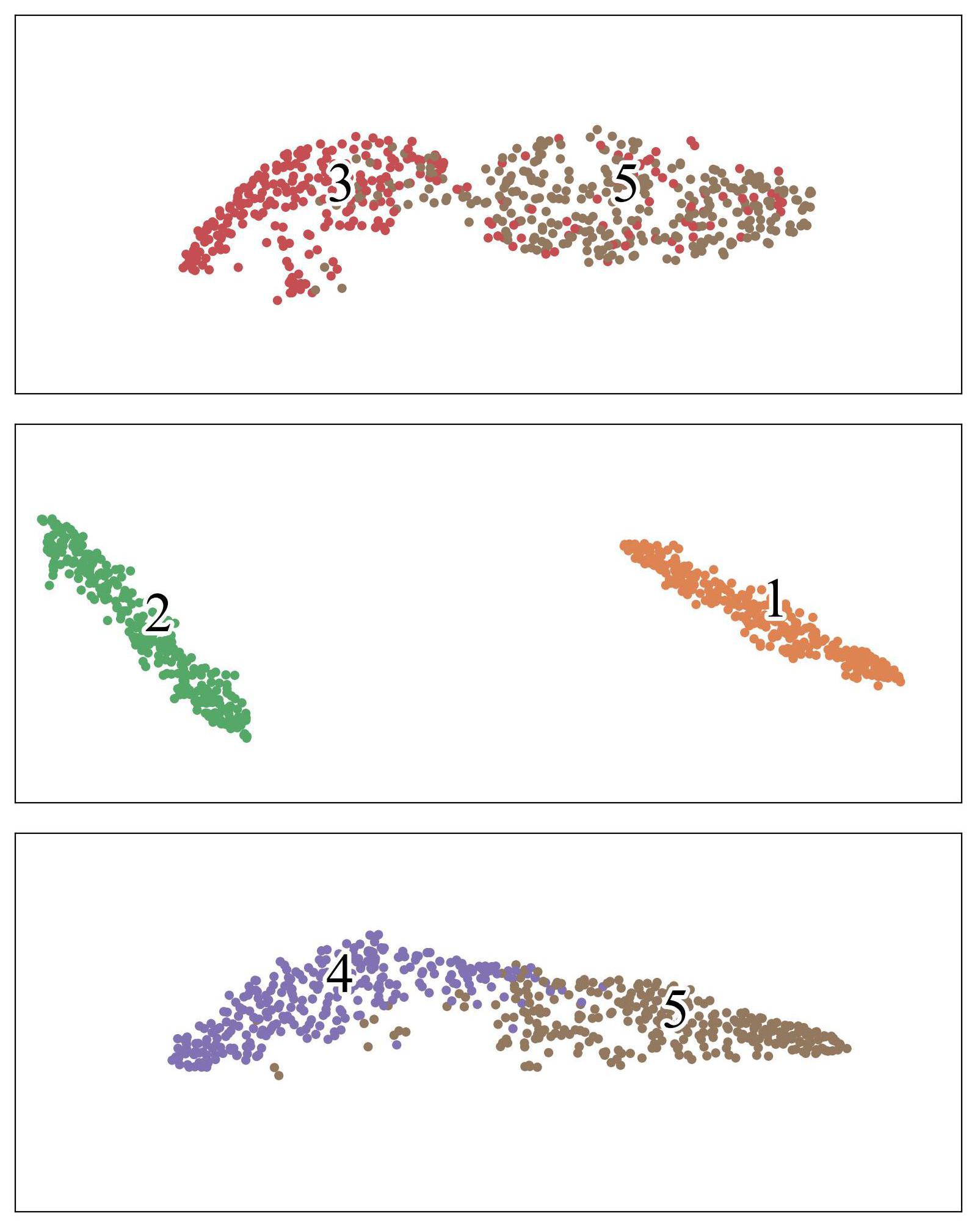}}
	\subfloat[]{\includegraphics[height=0.23\textheight]{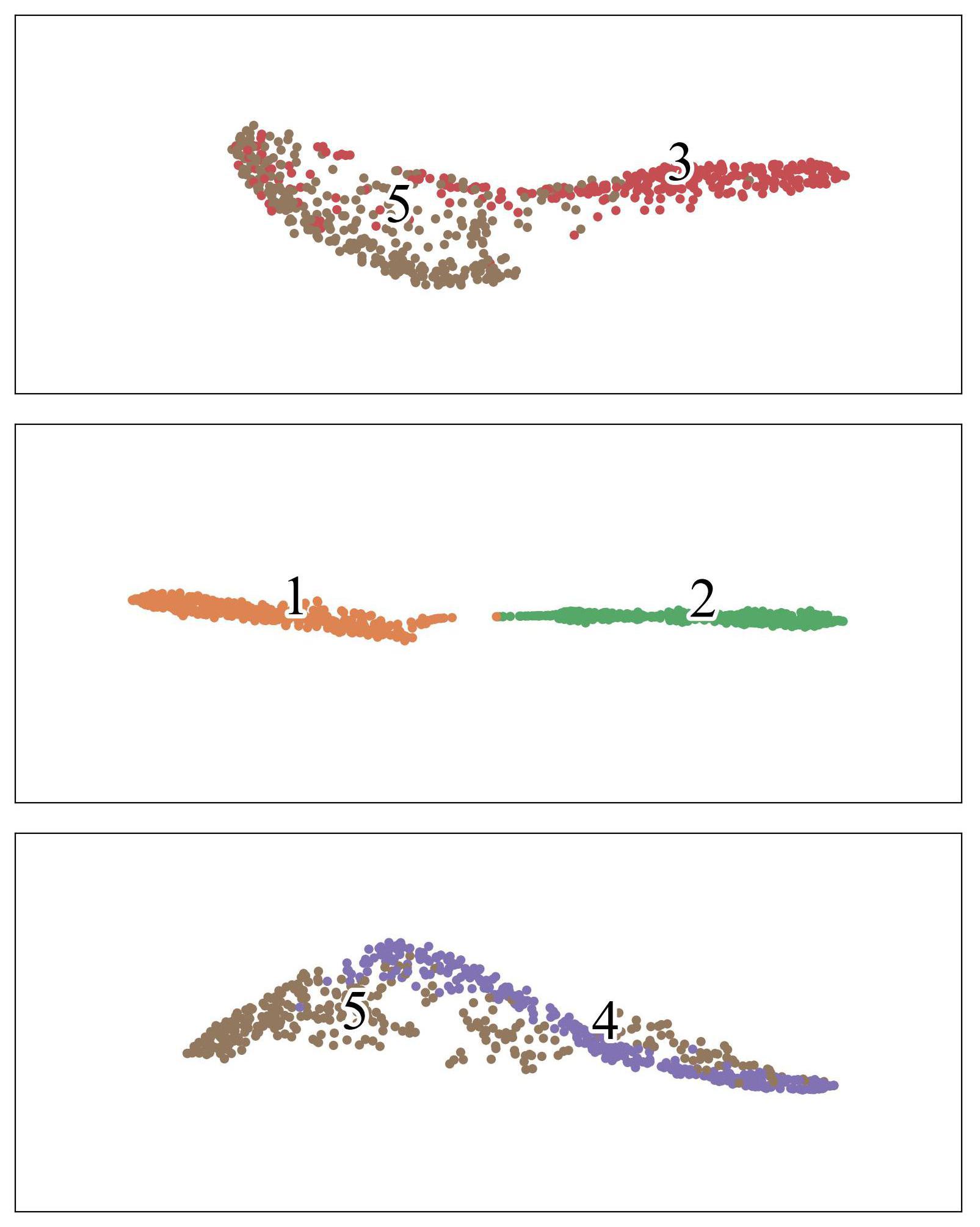}}
	\subfloat[]{\includegraphics[height=0.23\textheight]{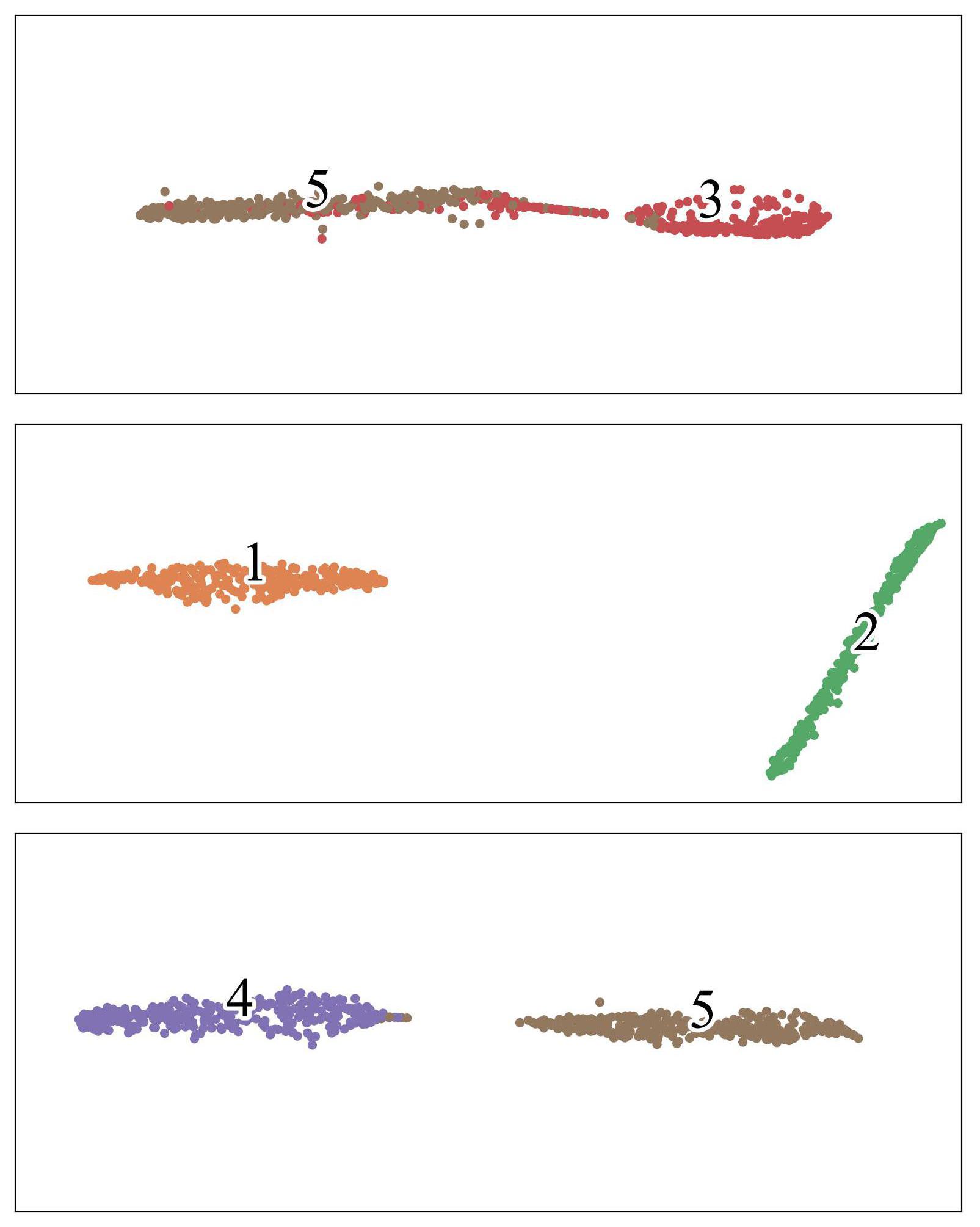}}\\
	\caption{Visualization of extracted representations using t-SNE. (a) FedAvg. (b) FedRep. (c) FedALA. (d) AFedCL.}
	\label{fig.7}
	\vspace{0.1cm}

\end{figure*}

\begin{table*}[t]
	\centering
	\vspace{0.2cm}
	\caption{Experimental results for different Non-IID settings and training sample sizes}
	\label{tab:optimized_results}
	\setlength{\tabcolsep}{5pt}
	\begin{tabular}{ c|c|cc|cc|cc|cc|cc|cc }
		\toprule
		& & \multicolumn{6}{c|}{Disjoint} & \multicolumn{6}{c}{Dirichlet} \\
		\multirow{1}{*}{Dataset} & \multirow{1}{*}{Methods} & \multicolumn{2}{c|}{5} & \multicolumn{2}{c|}{10} & \multicolumn{2}{c|}{20} & \multicolumn{2}{c|}{5} & \multicolumn{2}{c|}{10} & \multicolumn{2}{c}{20} \\
		& & Acc & F1 & Acc & F1 & Acc & F1 & Acc & F1 & Acc & F1 & Acc & F1 \\
		\midrule
		\multirow{9}{*}{NEU-CLS} 
		& No-FL    & 22.93 & 19.13 & 43.07 & 42.19 & 47.33 & 45.98 & 14.78 & 12.57 & 21.39 & 20.82 & 20.17 & 19.85 \\
		& FedAvg   & 51.40 & 62.83 & 58.60 & 70.89 & 72.47 & 80.86 & 44.70 & 51.50 & 57.39 & 64.04 & 60.01 & 67.68 \\
		& FedProx  & 51.53 & 61.42 & 61.93 & 73.96 & 73.80 & 82.22 & 42.78 & 47.03 & 52.69 & 60.76 & 64.87 & 71.51 \\
		& FedPer \cite{arivazhagan2019federated}   & 69.60 & 63.61 & 90.13 & 89.96 & 93.67 & 93.63 & 72.87 & 65.55 & 83.65 & 79.56 & 85.92 & 80.65 \\
		& FedRep \cite{collins2021exploiting}   & 74.73 & 68.87 & 87.87 & 87.28 & 93.33 & 93.31 & 70.78 & 62.01 & 81.57 & 75.74 & 86.96 & 80.79 \\
		& Ditto \cite{li2021ditto}    & 69.93 & 61.47 & 91.33 & 91.17 & 95.73 & 95.73 & 71.13 & 62.59 & 83.65 & 79.65 & 88.70 & 84.74 \\
		& FedALA \cite{zhang2023fedala}   & 76.46 & 70.95 & 90.27 & 89.93 & 93.33 & 93.46 & 73.56 & \textbf{67.61} & 82.44 & 76.91 & 86.09 & 82.67 \\
		& \textbf{Proposed AFedCL} & \textbf{78.67} & \textbf{74.84} & \textbf{95.20} & \textbf{95.19} & \textbf{96.60} & \textbf{96.60} & \textbf{73.91} & 67.27 & \textbf{86.96} & \textbf{83.77} & \textbf{88.87} & \textbf{85.32} \\
		
		\midrule
		
		\multirow{9}{*}{X-SDD}  
		& No-FL    & 31.03 & 28.16 & 35.49 & 33.32 & 39.15 & 35.82 & 23.56 & 20.69 & 25.32 & 24.50 & 26.01 & 25.03 \\
		& FedAvg  & 33.25 & 35.79 & 40.82 & 49.73 & 48.49 & 57.84 & 52.03 & 55.91 & 54.59 & 59.84  & 64.61 & 69.62   \\
		& FedProx & 37.02 & 39.97 & 42.36 & 52.24  & 50.80 & 61.32     & 52.22 & 55.96 & 55.32 & 61.67  & 61.84 & 68.32   \\
		& FedPer \cite{arivazhagan2019federated}   & 45.86 & 36.82 & 56.25 & 52.69  & 64.99 & 64.06     & 84.38 & 80.94 & 85.71 & 82.92  & 88.54 & 86.60   \\
		& FedRep \cite{collins2021exploiting}    & 44.52 & 31.56 & 54.08 & 49.13  & 66.73 & 65.79     & 83.57 & 79.39 & 84.84 & 81.58  & 88.88 & 87.76   \\
		& Ditto \cite{li2021ditto}        & 48.69 & 39.18 & 56.66 & 50.02  & 71.01 & 69.58     & 84.78 & 83.04 & 85.40 & 81.32  & 88.28 & 86.50   \\
		& FedALA \cite{zhang2023fedala}   & 54.14 & 48.03 & 60.68 & 57.79  & 67.45 & 65.22     & 85.96 & 82.70 & 86.27 & 83.90  & 87.62 & 86.21   \\
		& \textbf{Proposed AFedCL}                         & \textbf{58.17} & \textbf{53.60} & \textbf{64.89} & \textbf{62.53}  & \textbf{73.07} & \textbf{72.34} 
			& \textbf{87.02} & \textbf{83.15} & \textbf{89.27} & \textbf{88.31}  & \textbf{89.61} & \textbf{88.41} \\

			\midrule

			\multirow{9}{*}{GC10-DET} 
			& No-FL   & 22.82 & 18.36 & 34.04 & 32.85  & 36.25 & 34.31     & 18.43 & 16.37 & 17.04 & 15.34  & 30.48 & 32.17   \\
			& FedAvg                                       & 44.89 & 44.77 & 55.70 & 57.88  & 67.65 & 70.66     & 53.53 & 61.49 & 54.67 & 63.25  & 65.73 & 71.70   \\
			& FedProx                                      & 47.42 & 46.49 & 59.64 & 60.42  & 67.63 & 69.43     & 55.32 & 62.56 & 60.53 & 67.96  & 68.15 & 73.37   \\
		    & FedPer \cite{arivazhagan2019federated}                                       & 46.13 & 35.31 & 63.53 & 60.32  & 73.10 & 71.03     & 74.86 & 68.40 & 78.86 & 78.24  & 82.98 & 82.78   \\
			& FedRep \cite{collins2021exploiting}                                       & 45.90 & 35.48 & 59.96 & 55.75  & 73.15 & 70.77     & 71.54 & 64.53 & 79.54 & 76.44  & 84.28 & 82.26   \\
			& Ditto \cite{li2021ditto}                                           & 45.65 & 35.91 & 62.61 & 57.24  & 75.70 & 74.17     & 72.10 & 65.77 & 84.11 & 80.81  & 85.68 & 82.66   \\
			& FedALA \cite{zhang2023fedala}                                    & 50.99 & 44.16 & 64.56 & 60.80  & 71.70 & 70.73     & 78.19 & 72.27 & 80.44 & 77.24  & 82.56 & 81.30   \\
			& \textbf{Proposed AFedCL}                      &  \textbf{56.67} &  \textbf{48.52} & \textbf{70.18} & \textbf{68.47} & \textbf{77.10} & \textbf{75.84} 
			&  \textbf{80.46} &  \textbf{75.74} & \textbf{84.63} & \textbf{81.70}  & \textbf{87.07} & \textbf{85.35} \\

		\bottomrule
		
	\end{tabular}
	\vspace{0.1cm}
\end{table*}

	\subsection{Visual Analyses}
	To intuitively compare model performance across methods, we utilize the t-Distributed Stochastic Neighbor Embedding (t-SNE) technique to visualize the feature representations extracted by FedAvg, FedRep, FedALA, and our AFedCL method using the NEU-CLS dataset in a disjoint setting where each client has two random categories. In Fig. 7, rows and columns represent clients and methods, respectively, with different colors indicating distinct data categories. 
	It demonstrates that AFedCL achieves more cohesive clustering within classes and clearer separation between different classes, particularly in the tasks of client 0 and client 2. In contrast, other methods struggle to clearly distinguish between different classes in these cases, highlighting the enhanced performance of the proposed AFedCL method.\par

	\subsection{Comparison With Benchmark Methods}
	In this part, we evaluate the proposed AFedCL method under two non-IID settings and varying numbers of available samples in each client's local training. Table \MakeUppercase{\romannumeral 2} presents the testing results of different methods on three datasets. 
	For No-FL, where each client learns only from individual local data, the performance is the worst, reflecting the problem posed by data scarcity and the necessity of adopting FL.
	Among the FL methods, the two one-fit-all methods, including FedAvg and FedProx, generally exhibit inferior performance. This can be attributed to the challenge posed by data heterogeneity, impairing the performance of the global model. \par

	Meanwhile, other PFL methods, including FedPer, FedRep, Ditto, and FedALA, perform better since they retain personalized knowledge while absorbing global knowledge from the federation. 
	In particular, benefiting from the adversarial training strategy designed to overcome the global knowledge forgetting problem, along with the feature fusion module, our proposed AFedCL method attains superior performance across different heterogeneity settings and varying training sample numbers in all three datasets. It provides an increase in accuracy by up to 3.87\%, 4.21\%, and 5.67\% on the three datasets, respectively, highlighting the method's effectiveness in mitigating the challenge of data heterogeneity and improving overall model performance. \par

	\subsection{Sensitivity Analyses}
	Here, we study the robustness of the proposed method under various factors: the balanced weight of the discrimination loss function $\lambda$, the backbone of the model, and the number of participating clients $K$. \par

	Firstly, we explore the impact of varying $\lambda$ on the testing performance. Throughout all experiments, $\lambda$ is tuned by selecting the best-performing value from a candidate set $\{0.01, 0.1, 1\}$ for both our method and a competitive baseline, Ditto, which also employs a $\lambda$ to achieve a balance between two loss functions. As shown in Fig. 8, the results reveal that AFedCL consistently outperforms Ditto in testing accuracy and exhibits smaller fluctuation across different $\lambda$ values, demonstrating its stability and superiority. \par
	
	The impact of changing the model's backbone from MobileNet to ResNet-18 and ResNet-50 is also explored. Results in Fig. 9(a) reveal that AFedCL exceeds other methods with each backbone, highlighting its robustness to backbone variations. Finally, we vary the number of clients while maintaining a constant total of 50 training samples, distributed as 50 samples for 2, 5, and 10 clients, respectively. As indicated in Fig. 9(b), AFedCL shows superior performance across varying client numbers, proving its robustness to changes in client participation. Overall, these analyses underline the proposed method's adaptability and reliability under diverse conditions. \par
	
	\begin{figure}[t]

	\centering

	\subfloat[]{\includegraphics[width=0.49\columnwidth]{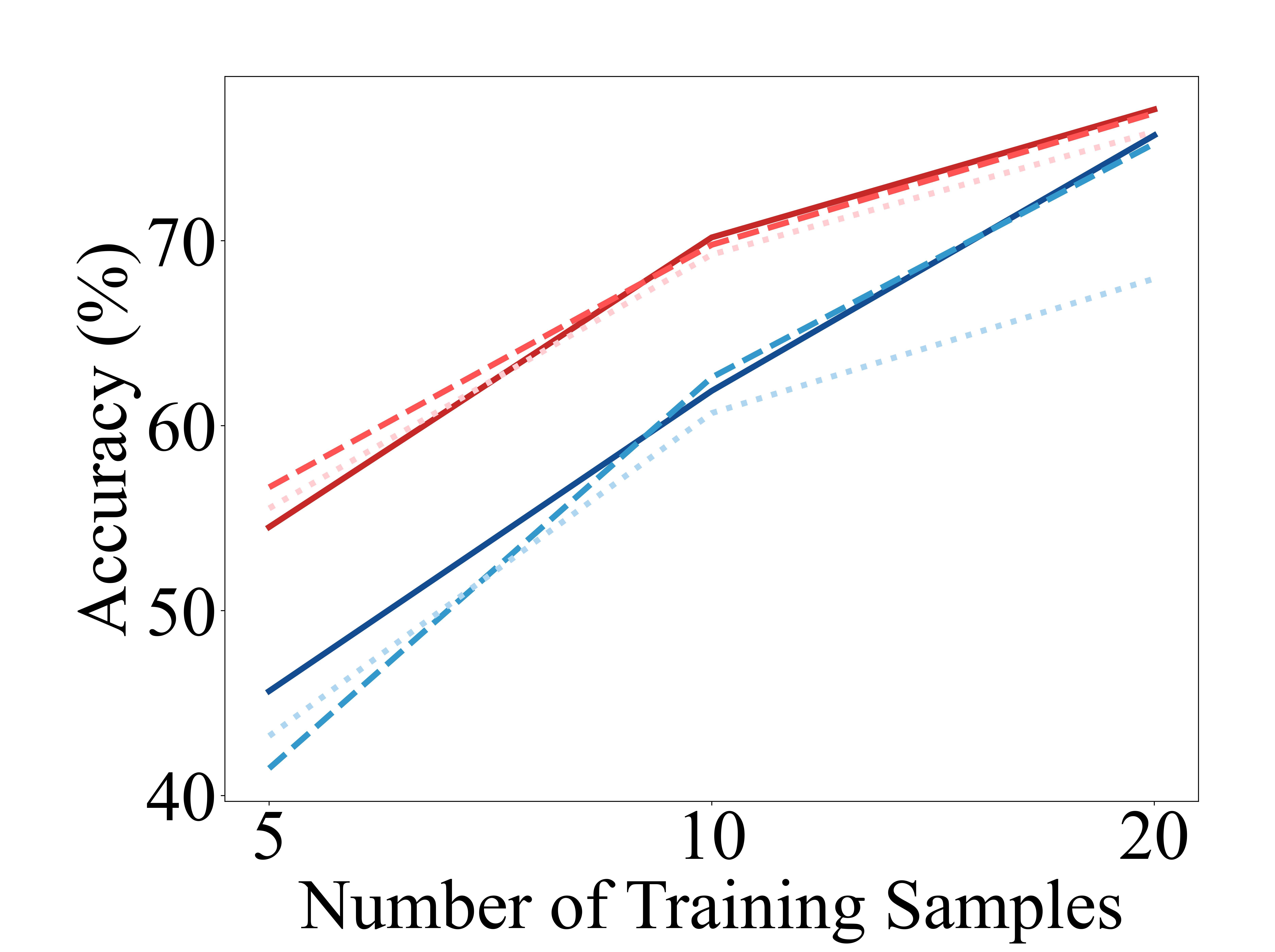}}\hspace{-0.2cm}
	\subfloat[]{\includegraphics[width=0.49\columnwidth]{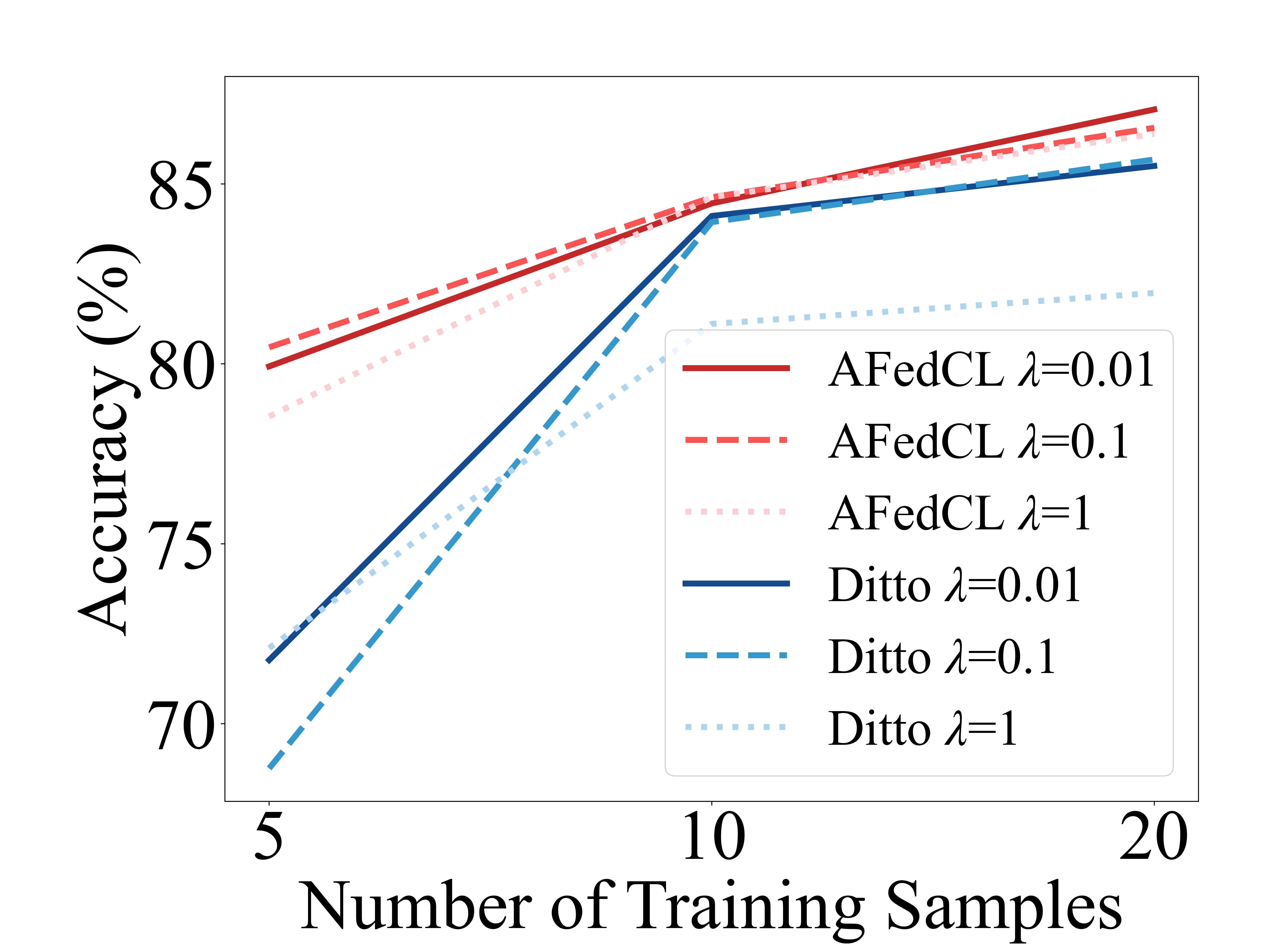} } \\
	
	\caption{Performance comparison of AFedCL and Ditto for various $\lambda$ values. (a) Disjoint setting. (b) Dirichlet setting.}
	\label{fig.8}

\end{figure}
	\begin{figure}[t]
	\centering
	\subfloat[]{\includegraphics[height=0.12\textheight]{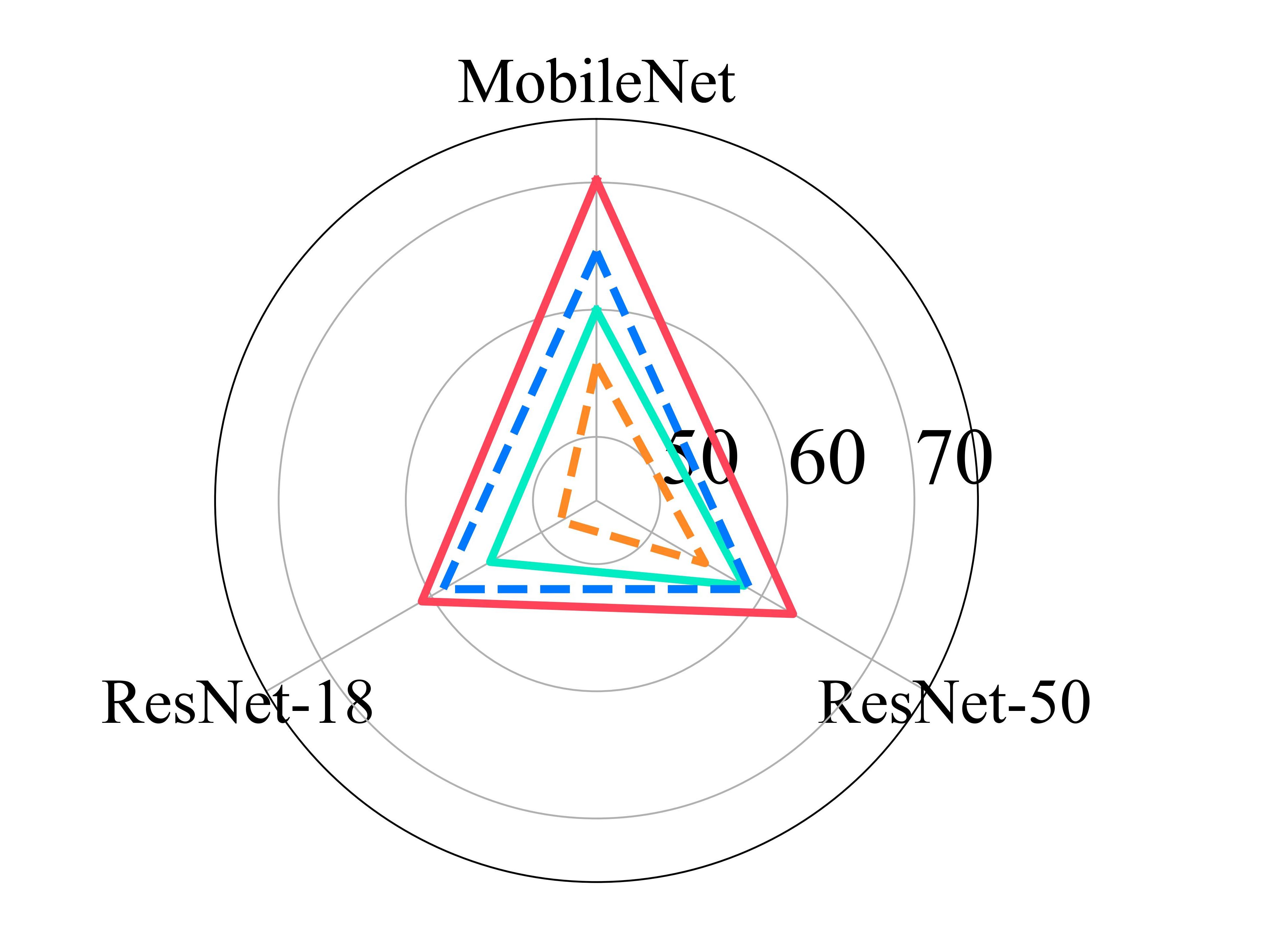}}\hspace{-0.2cm}
	\subfloat[]{\includegraphics[height=0.12\textheight]{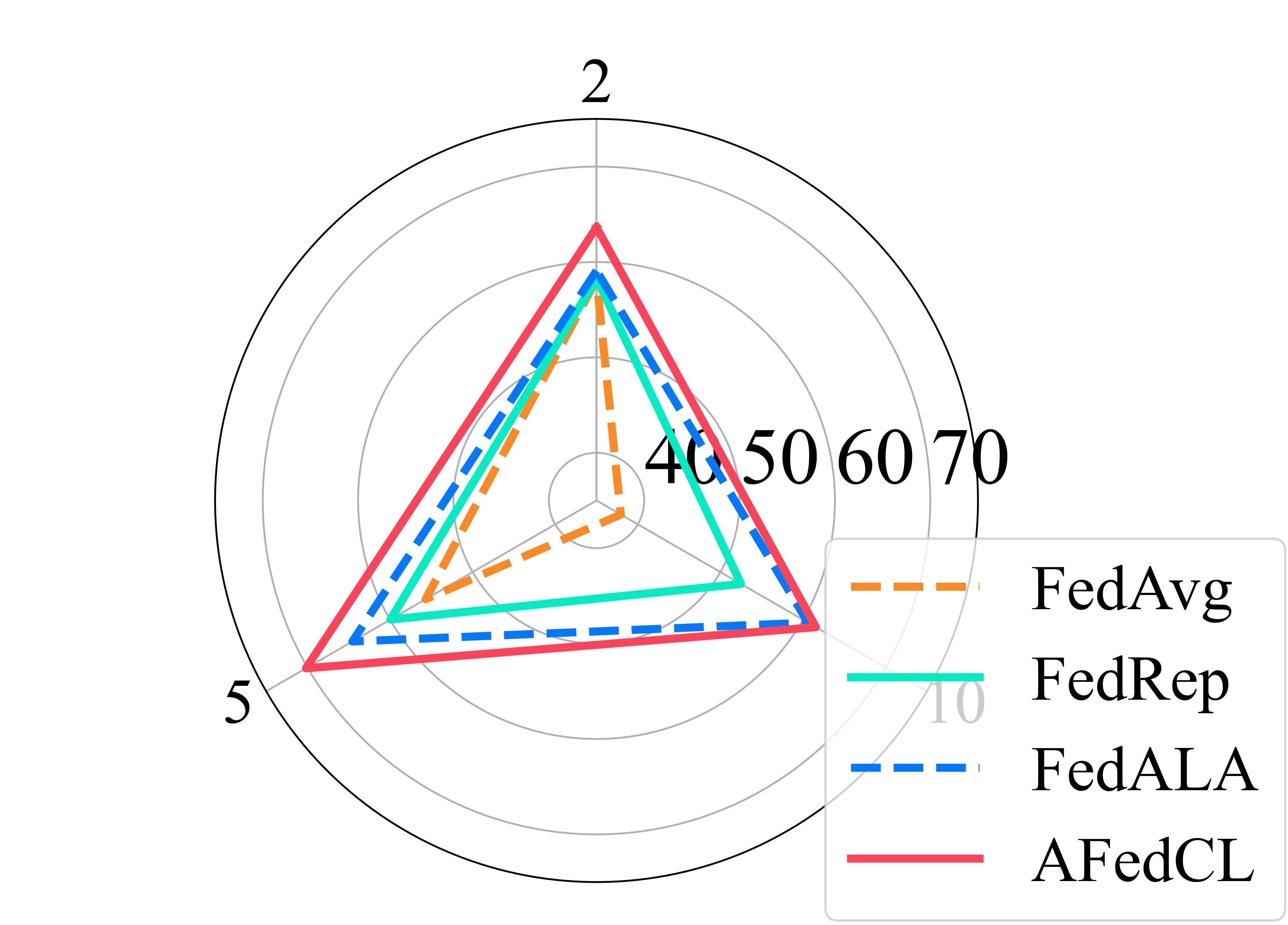} } \\
	\caption{Performance across various backbones and numbers of clients. (a) different backbones. (b) different numbers of clients.}
	\label{fig.9}
\end{figure}
	
	\subsection{Ablation Experiments}
	To confirm each component in the AFedCL contributes to performance, including the dynamic consensus construction (DCC) strategy, the consensus-aware aggregation (CAA) mechanism, and the adaptive feature fusion (AFF) module, the ablation experiments are conducted using the GC10-DET dataset across two non-IID settings and with two training sample sizes. The results, illustrated in Fig. 10, show a stepwise enhancement in performance across these variants up to the complete method, highlighting the concrete impact of each component on the method's efficacy.
	
	\begin{figure}[t]
	\centering

	\subfloat[]{\includegraphics[width=0.49\columnwidth]{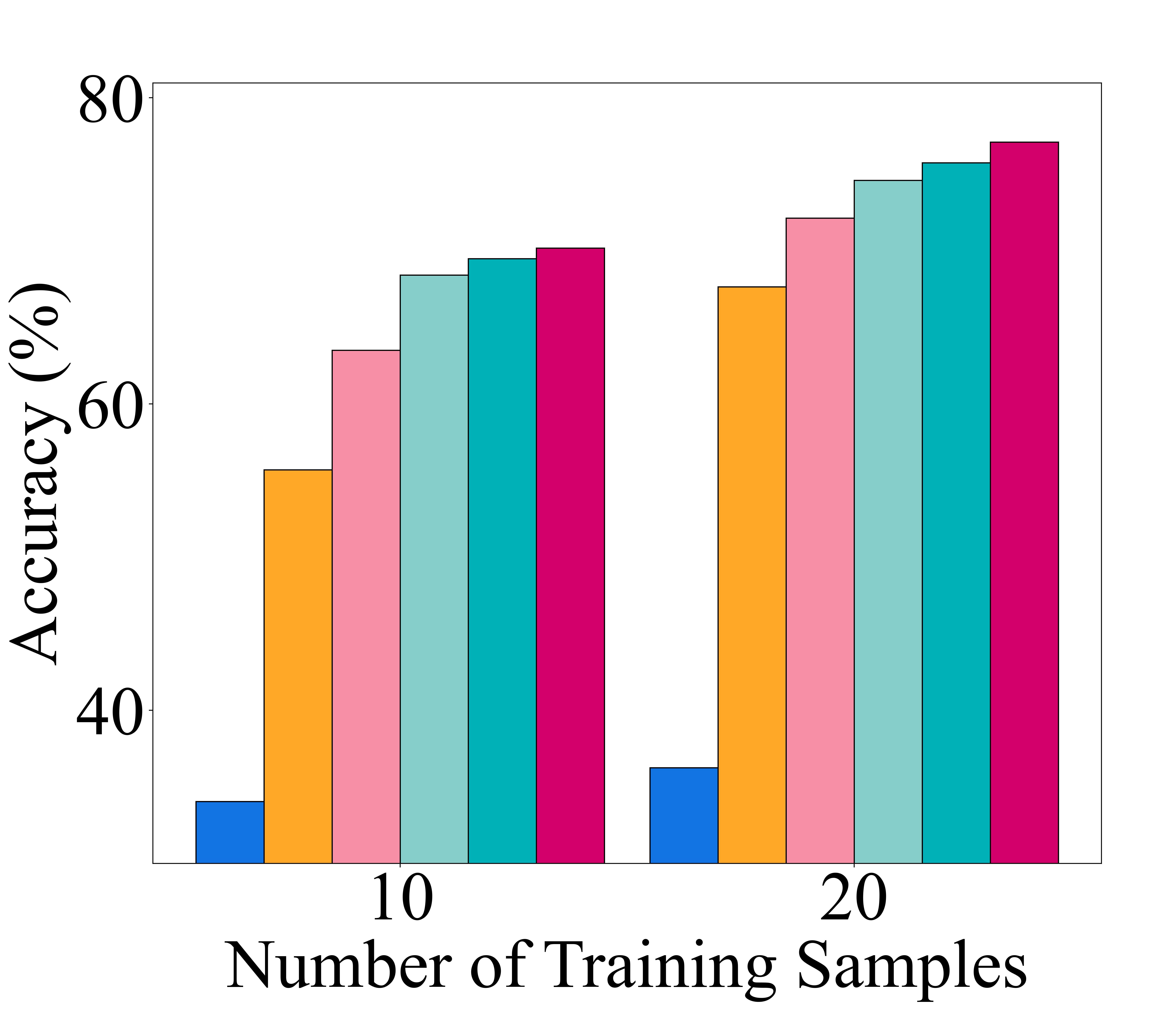}}\hspace{-0.2cm}
	\subfloat[]{\includegraphics[width=0.49\columnwidth]{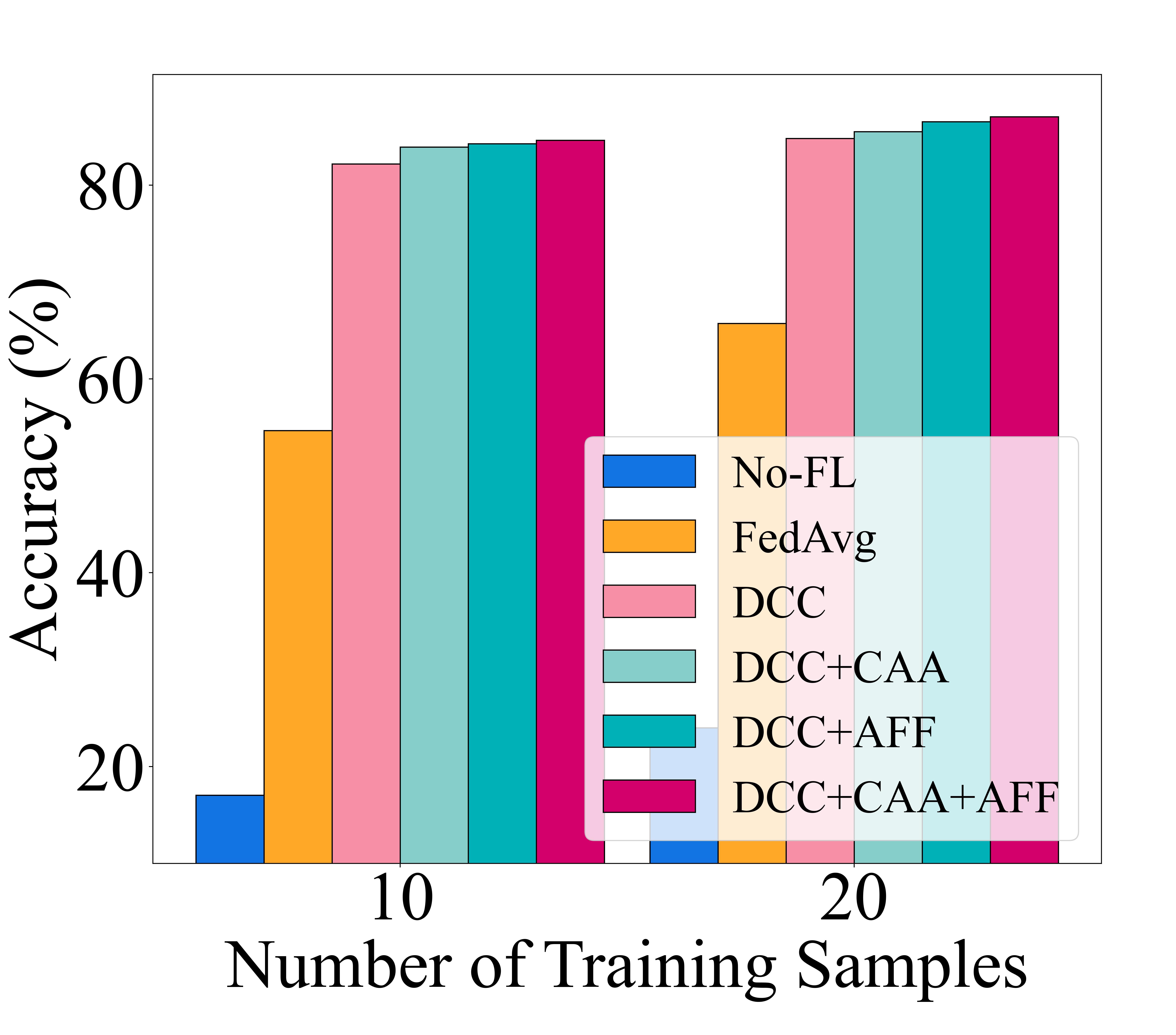} } \\
	
	\caption{Results of ablation experiments. (a) Disjoint setting. (b) Dirichlet setting.}
	\label{fig.10}
\end{figure}

	% 【6】
	\section{Conclusion}
	In this paper, we have developed a novel PFL approach, AFedCL, to tackle the challenge of data heterogeneity in SDC. It enables multiple industrial entities in IIoT to collaboratively train personalized models while preserving privacy. 
	Specifically, a dynamic consensus construction strategy has been invented to achieve distribution alignment among different clients through an adversarial game process, thus mitigating performance degradation caused by global knowledge forgetting under non-IID data. 
	Building on this, a consensus-aware aggregation mechanism has been introduced. It assigns aggregating weights based on the global knowledge learning efficacy, measured by each client's discrimination loss, improving the generalization ability of the global model.
	Furthermore, an adaptive feature fusion module has been designed. Through the optimization of personalized scalar fusion weights of global and local features, the utilization efficiency of global knowledge is enhanced, thus boosting model performance.
	The proposed method demonstrates promising results across extensive experiments on three real-world SDC datasets, achieving a notable increase in accuracy of up to 5.67\%. \par
	
	\bibliographystyle{IEEEtran}
	\bibliography{myreference}
	
\end{document}